\documentclass[preprint,12pt]{elsarticle}
\pdfoutput=1
\usepackage[fleqn]{amsmath} 
\usepackage{amssymb}
\usepackage{graphicx}
\usepackage{subfig} 
\usepackage{float} 
\usepackage[numbers]{natbib}
\usepackage{tcolorbox}
\usepackage{rotating}
\usepackage{url}
\usepackage{hyperref}
\usepackage{stfloats}
\usepackage{longtable}
\usepackage{booktabs}
\usepackage[space]{grffile}
\usepackage{multirow}
\usepackage[font=small,labelfont=bf]{caption}
\usepackage{makecell}
\usepackage{lineno}
\usepackage{adjustbox}

\newcommand{\etal}{{et al}.~}
\newcommand{\dotp}[2]{\ensuremath{\langle #1 , #2 \rangle}}

\def\defeq{\mathrel{\mathop:}=}

\journal{arXiv}
\begin{document} 

\begin{frontmatter}

\title{The Trace Criterion for Kernel Bandwidth Selection for Support Vector Data Description}
\author[sas]{Arin Chaudhuri\corref{cor1}}
\ead{arin.chaudhuri@sas.com}
\author[sas]{Carol Sadek}
\author[sas]{Deovrat Kakde}
\author[goog]{Wenhao Hu\fnref{wenhao}}
\author[sas]{Hansi Jiang}
\author[sas]{Seunghyun Kong}
\author[sas]{Yuwei Liao}
\author[sas]{Sergiy Peredriy}
\author[sas]{Haoyu Wang}
\cortext[cor1]{Corresponding author}
\address[sas]{Internet of Things, SAS Institute Inc., Cary, NC, 27513}
\address[goog]{2250 Latham St Apt 51, Mountain View, CA 94040}
\fntext[wenhao]{This work was done when the author was with SAS Institute Inc.}

\begin{abstract} 
Support vector data description (SVDD) is a popular anomaly detection 
technique. The SVDD classifier partitions the whole data space
into an \emph{inlier} region, which consists of the region \emph{near} the training
data, and an \emph{outlier} region, which consists of points \emph{away} from the
training data. The computation of the SVDD classifier requires a kernel function,
for which the Gaussian kernel is a common choice. The Gaussian
kernel has a bandwidth parameter, and it is important to set the value of this
parameter correctly for good results. A small bandwidth leads to overfitting such that
the resulting SVDD classifier overestimates the number of anomalies, whereas a
large bandwidth leads to underfitting and an inability to detect many anomalies. In this paper,
we present a new unsupervised method for selecting
the Gaussian kernel bandwidth. Our method exploits a low-rank representation of 
the kernel matrix to suggest a kernel bandwidth value. Our new technique is competitive 
with the current state of the art for low-dimensional data and performs extremely well for many 
classes of high-dimensional data. Because the mathematical formulation of SVDD is identical
with the mathematical formulation of one-class support vector machines (OCSVM) when the 
Gaussian kernel is used, our method is equally applicable to Gaussian kernel bandwidth tuning
for OCSVM.
\end{abstract}


\end{frontmatter}
\section{Introduction}
\addtocontents{toc}{\protect\setcounter{tocdepth}{2}}
\label{intro}
In this section we provide a brief description of the support vector data
description technique. Our description follows the description given in
\cite{tax2004support} and \cite{2016arXiv160205257K,ChaudhuriMean}.

Support vector data description (SVDD) is a machine learning technique that is
used for single-class classification and anomaly detection. First introduced
by Tax and Duin \cite{tax2004support}, SVDD's mathematical formulation is
almost identical to the one-class variant of support vector machines: one-class
support vector machines (OCSVM), which is attributed to Sch{\"ol}kopf \etal
\cite{scholkopf2000support}. The use of SVDD is popular in domains in which the
majority of data belong to a single class and no distributional assumptions can
be made. For example, SVDD is useful for analyzing sensor readings from reliable
equipment for which almost all the readings describe the equipment's normal
state of operation.

SVDD provides a geometric description of the observed data. The SVDD classifier
assigns a \emph{distance} to each point in the domain space; the distance
measures the separation of that point from the training data. During scoring,
any observation found to be at a \emph{large} distance from the training data
might be an anomaly, and the user might choose to generate an alert.

Several researchers have proposed using SVDD for multivariate process control
\cite{sukchotrat2009one,camci2008general}. Other applications of SVDD involve
monitoring condition of machines \cite{widodo2007support,ypma1999robust} and image
classification \cite{sanchez2007one,liao2018new}.

\subsection{Mathematical Formulation}
In this section, we describe the mathematical formulation of SVDD
following \cite{tax2004support,ChaudhuriMean}.

\subsubsection{Normal Data Description}
The SVDD model for normal data description builds a sphere that contains
most of the data within a small radius. Given observations $x_1,\dots,x_n \in {\mathbb{R}}^{p}$,
we need to solve an optimization problem to determine the sphere. We present the 
optimization problem in its primal and dual forms below.
\newpage
\noindent
{\bf Primal Form}\par
\ \newline
\scalebox{0.85}{
    \fbox{
        \parbox{\columnwidth}{
            \ \newline
            \noindent{\bf Objective}
            \begin{equation}
                \mathrm{min}~ R^{2} + C\sum_{i=1}^{n}\xi_{i}, 
            \end{equation}
            {\bf Subject to} 
            \begin{align}
                \|x_{i}-a\|^2 \leq R^{2} + \xi_{i}, \forall i=1,\dots,n,\\
                \xi_{i}\geq 0, \forall i=1,\dots,n,
            \end{align}

            {\bf Where}\newline
            \parbox{0.8\columnwidth}{
                \begin{itemize}
                    \item $x_{i} \in {\mathbb{R}}^{p}, i=1,\dots,n  $ represent the training data,
                    \item $R$ is the radius and represents the decision variable,
                    \item $\xi_{i}$ is the slack for each variable,
                    \item $a$ is the center (a decision variable),
                    \item $C$ is the penalty constant that controls the trade-off between 
                        the radius and the errors. 
                \end{itemize}
            }
        }
    }
}
\par
\ \newline
\noindent\textbf{Dual Form}\newline
\ \newline
The dual formulation is obtained using Lagrange multipliers.
\ \newline \ \newline
\scalebox{0.85}{
    \fbox{
        \parbox{\columnwidth}{
            \mbox{}\newline
            {\bf Objective}
            \begin{equation} 
                \textrm{max}~ \sum_{i=1}^{n}\alpha _{i}\dotp{x_{i}}{x_{i}} - 
                \sum_{i,j}^{ }\alpha _{i}\alpha _{j}\dotp{x_{i}}{x_{j}} ,
            \end{equation}
            {\bf Subject to}
            \begin{align*}
                \sum_{i=1}^{n}\alpha _{i} = 1,\\
                0 \leq  \alpha_{i}\leq C,\forall i=1,\dots,n
            \end{align*}
            \par
            \noindent\textbf{Where}\newline\mbox{}
            \parbox{\columnwidth}{
                $\alpha_{i}\in \mathbb{R}$ are the Lagrange constants and $C$ is the penalty constant.
            }
        }
    }
}
\newpage
\noindent{\bf Duality Information}

The optimal dual parameters $\alpha_i$ are extremely useful as they can be used to      
determine the minimal radius sphere which surrounds the data and the position of
the points in the data relative to this sphere.

Let $SV_{<~C}$ denote the set $\{ x_j : 0 < \alpha_j < C \}$ and choose any $x_k \in SV_{<~C}$.
The radius of the sphere $\mathbf{R}$  is given by
\begin{equation}   
    \mathbf{R}^{2}=\dotp{x_{k}}{x_{k}}-2\sum_{i}^{ }\alpha _{i}\dotp{x_{i}}{x_{k}}+\sum_{i,j}^{ }\alpha _{i}\alpha _{j}\dotp{x_{i}}{x_{j}}.
\end{equation}
The value of $\mathbf{R}^2$ does not depend on the choice of $x_k \in SV_{<~C}$.\par
The center of the sphere $\mathbf{a}$ is given by
\begin{equation}
    \mathbf{a} = \sum_{i=1}^{n}\alpha _{i}x_{i} 
\end{equation}

The points $x_i$ for which the corresponding $\alpha_i$ equal $0$ lie inside this sphere,
that is,
\begin{equation} \left \| x_{i}- \mathbf{a} \right \| < \mathbf{R} \iff \alpha _{i} = 0.\end{equation}

The points $x_i$ for which the corresponding $\alpha_i$ lie between $0$ and $C$ are 
on the boundary of this sphere. 
\begin{equation*} \left \| x_{i}- \mathbf{a} \right \| = \mathbf{R} \iff 0< \alpha _{i}< C\end{equation*}

The remaining points lie outside the sphere.
\begin{equation}\left \| x_{i}-\mathbf{a}\right \| > \mathbf{R} \iff \alpha _{i}= C\end{equation}

Any $x_i$ for which the corresponding $\alpha_i > 0$ is called a support vector.\par
\noindent{\bf Scoring}\par
For any point $z$,  its distance from the center of the sphere $ \textrm{dist}^{2}(z) = \|z - \mathbf{a}\|^2 $ 
equals
\begin{align}  
    \textrm{dist}^{2}(z) &= \| z - \sum_{i} \alpha_i x_i \|^2  \nonumber \\
                         &=  \langle z, z \rangle - 2\sum_{i}^{ }\alpha _{i}\langle x_{i}, z \rangle
    +\sum_{i,j}^{}\alpha_{i}\alpha_{j}\langle x_{i},x_{j}\rangle. 
\end{align}
 If $\textrm{dist}^{2}(z) > R^{2} $ then $z$ is designated as an outlier.\par
\noindent{\bf Problems with the Normal Data Description}\par 
The spherical data boundary can include a significant amount of space that
has a sparse distribution of training observations. Using this model to score
can lead to a lot of false positives. Hence, instead of a spherical shape, a
compact bounded outline around the data is often desired. Such an outline should
approximate the shape of the single-class training data. This is possible
by using kernel functions. The SVDD forumation with kernel functions leads to 
the flexible data description which is described next.
\subsubsection{Flexible Data Description}
The support vector data description is made flexible by replacing the inner
product $ \dotp{x_{i}}{x_{j}} $ with a suitable kernel function $ K(x_{i},x_{j})
$. The Gaussian kernel function used in this paper is defined as
\begin{equation}  
K_s(x, y)= \mathrm{exp}  \dfrac{ -\|x - y\|^2}{2s^2};
 \end{equation}
where $s$ is the Gaussian bandwidth parameter.

The modified mathematical formulation of SVDD with a kernel function $K$ is\par \mbox{}\newline
\scalebox{0.85}{
    \fbox{
        \parbox{\columnwidth}{
            \mbox{}\newline
            {\bf Objective}
            \begin{equation} 
                \textrm{max}~ \sum_{i=1}^{n}\alpha_{i} K(x_{i},x_{i}) - 
                \sum_{i,j}^{ }\alpha_{i}\alpha_{j}K(x_{i},x_{j}) ,
            \end{equation}
            {\bf Subject to}
            \begin{align*}
                \sum_{i=1}^{n}\alpha _{i} = 1,\\
                0 \leq  \alpha_{i}\leq C,\forall i=1,\dots,n
            \end{align*}
            {\bf Where} \newline
            \parbox{\columnwidth}{
                $\alpha_{i}\in \mathbb{R}$ are the Lagrange constants, and $C$ is the penalty constant.
            }
        }
    }
}
\par\mbox{}\\
In perfect analogy with the previous section, any $x_i$ for which
$\alpha_i = 0$ is an \emph{inside} point and any $x_i$ for
which $\alpha_i > 0$ is called a support vector.\par
$SV_{<~C}$ is similarly defined as $\{x_j : 0 < \alpha_j < C\}$ 
and given any $x_k \in SV_{<~C}$ the threshold $R^{2}$ is calculated as 
\begin{equation}
R^{2} = K(x_{k},x_{k})-2\sum_{i}^{ }\alpha _{i}K(x_{i},x_{k})+\sum_{i,j}^{ }\alpha _{i}\alpha _{j}K(x_{i},x_{k})
\end{equation}
The value of $R^2$ does not depend on which $x_k \in SV_{<~C}$ is used.\par
\noindent{\bf Scoring}\par
For any observation $z$, the distance $\text{dist}^{2}(z)
$ is calculated as follows: 
\begin{equation}
\text{dist}^{2}(z)= K(z,z) - 2\sum_{i}^{ }\alpha_{i}K(x_{i},z)+
\sum_{i,j}^{ }\alpha_{i}\alpha_{j}K(x_{i},x_{j})
\end{equation}
Any point $z$ for which $\text{dist}^{2}(z) > R^{2} $ is designated as an outlier.
\subsection{Overview}
\label{sec:imp}
In practice, SVDD is almost always computed by using the Gaussian kernel
function, and it is important to set the value of bandwidth parameter correctly.
A small bandwidth leads to overfitting, and the resulting SVDD classifier
overestimates the number of anomalies. A large bandwidth leads to underfitting,
and the classifier fails to detect many anomalies.

In this paper, we introduce the trace criterion method for Gaussian bandwidth
selection. The computation of the trace criterion consists of finding the
inflection point of a smooth function of the bandwidth parameter. The
computation terminates quickly when standard nonlinear optimization methods such
as Newton-Raphson are used. This method is efficient for moderately large data
sets: we have successfully applied the trace criterion method to datasets with
two million observations. The trace criterion has also been available in commerical 
software since May 2019~\cite{SASVDMMLTRACE}.

Our results show that the trace criterion is competitive with the current state
of the art, it can be computed quickly and simulation studies show that this method
is very accurate for high-dimensional data.

These properties make the trace criterion method a good bandwidth selection
technique. However, unsupervised bandwidth tuning is an extremely difficult
problem, so it is quite possible that there is a class of data sets for which
the trace criterion method does not give good results.

The rest of the paper is organized as follows. Section~\ref{sec:tcrit}
defines the trace criterion method for bandwidth tuning; in section~\ref{sec:trealt},
we introduce alternative bandwidth selection methods; and in 
section~\ref{sec:treval} we compare the different bandwidth selection methods on
a variety of simulated and real data.

\section{The Trace Criterion for Bandwidth Selection} \label{sec:tcrit}
\subsection{Parameter Tuning Using Inflection Points}
Parameter tuning for unsupervised learning methods such as clustering and
one-class classification can be difficult when external validation data
are not available, as is quite frequently the case. A popular method for
parameter tuning in such cases is to look at the values of a ``validation
measure'' as a function of the hyperparameter of interest, and choose the
value of the hyperparameter where the function has an inflection point
\citep{aggarwal2015data}. For a specific example, consider the problem of
determining the number of clusters for $K$-means. $SSQ$ (the sum of the squared
distance of each point in the training data from the cluster center closest to
it) can be taken as a validation measure. If the number of clusters is held
fixed, then a lower value of $SSQ$ indicates better clustering. Let $SSQ(K)$
denote the value of $SSQ$ for the clustering that is suggested by $K$-means
with $K$ clusters. As $K$ increases, $SSQ(K)$ decreases, so you cannot choose
the number of clusters as $K = \operatorname{argmin}_{K} SSQ(K)$ because doing
so would suggest as many clusters as the number of points in the training
data. However, it is observed that the value of $K$ at which the function $K
\to SSQ(K)$ has an inflection point is quite frequently a good choice for the
number of clusters. The inflection point of other validation measures such as
the silhouette coefficient is also used to determine the number of clusters
\citep{aggarwal2015data}.

We will now propose a validation measure for SVDD whose inflection point
provides our suggested bandwidth.

\subsection{Validation Measure for Trace Criterion}

Assume we have a training data set $x_1,\dots,x_N$ that consists of $N$ distinct
points in $\mathbb{R}^p$ and we propose the trace criterion as a kernel bandwidth value
for training this data set. The trace criterion uses the Nystr\"{o}m approximation
of a kernel matrix, so we start by describing the Nystr\"{o}m approximation.\par
\subsubsection{Nystr\"{o}m Approximation of the Kernel Matrix}\par
In kernel methods, including SVDD, the objective function to be optimized
depends on the training data through the kernel matrix $K=(K(x_i,x_j)).$ 
Because this matrix has $N^2$ elements, where $N$ is the
number of observations, it is impossible to work with the entire kernel
matrix even for moderate values of $N$. The Gaussian kernel matrix is always
positive semidefinite, and it typically has a rapidly decaying spectrum. The
rapidly decaying spectrum can be exploited to create a low-rank positive
semidefinite approximation $\tilde{K}$ of $K$ by replacing all the eigenvalues
in the spectral decomposition of $K$ below a certain threshold with $0$
\citep{NIPS2000_1866}. $\tilde{K}$ has a square root of low rank, that is,
$\tilde{K} = GG^T$, where $G$ is an $N \times r$ matrix with $r \ll N$. Since $K
\approx \tilde{K}$, $G$ is considered to be an approximate low-rank square root
of $K$. In many cases, computation of expressions that involve $K$ can be made
tractable by replacing $K$ with $GG^T$. However, it is not feasible to compute
$G$ such that $K \approx GG^T$ by using the eigendecomposition of $K$ when $N$
is large.\par
The Nystr\"{o}m methods form a class of popular methods to compute a low-rank
representation of $K$ even when $N$ is large. We use a variant of the Nystr\"{o}m
method as described in \citep{zhang2008improved} to construct a validation
measure whose inflection point will be suggested as the bandwidth.

Given data $x_1,\dots,x_N$ in $\mathbb{R}^p$ and an integer $0 < r \ll N$,
let $z_2,z_2,\dots,z_r$ be distinct \emph{landmark points} in $\mathbb{R}^p$.
The landmark points are chosen so that they are evenly distributed within the
training data. Following \citep{zhang2008improved}, we choose $z_1,\dots,z_r$
as the cluster centers that are obtained from a $K-$means clustering with $r$
clusters of the training data.

From the theory of reproducing kernel Hilbert spaces (RKHS), we know that for
any $s > 0$ there is a Hilbert space $\mathcal{H}_s$ and a mapping $\Phi \colon
\mathbb{R}^p \to \mathcal{H}_s$ such that for any $x,y \in \mathbb{R}^p$ we have
$\dotp{\Phi(x)}{\Phi(y)}_s = \exp-\dfrac{\|x-y\|^2}{2s^2} \defeq K_s(x,y)$,
where $\dotp{~}{~}_s$ denotes the inner product on $\mathcal{H}_s$ and $\| \|_s$
denotes the norm on $\mathcal{H}_s$.

Let $V_{0} \subseteq \mathcal{H}_s$ denote the subspace spanned by
$\Phi(z_1),\dots,\Phi(z_r)$. Note that $V_0$ is a finite-dimensional and hence
closed subspace of $\mathcal{H}_s$ of dimension exactly $r$. Given $x \in
\mathbb{R}^p$, let $\widehat{\Phi}(x)$ denote the projection of $\Phi(x)$ on
$V_{0}.$ In the Nystr\"{o}m methods, the kernel matrix
$\left(K_s(x_i,x_j)\right)_{i,j=1}^{N}=\left(\dotp{\Phi(x_i)}{\Phi(x_j)}\right)_
{i,j=1}^{N}$ is approximated by
$(\dotp{\widehat{\Phi}(x_i)}{\widehat{\Phi}(x_j)})_{i,j=1}^{N}$, a rank $r$
matrix that has a low-dimensional square root that can be easily computed.

\subsubsection{Trace Criterion validation measure}\par
We use the projected values $\widehat{\Phi}(x)$ for a different purpose. We
use the discrepancy between $\widehat{\Phi}(x)$ and ${\Phi}(x)$ for $x$ in the
training data to come up with a validation measure.

From usual least square arguments, we know that for any $x \in \mathbb{R}^p$,
we have $\widehat{\Phi}(x) = \sum_{i=1}^{r} \gamma_i \Phi(z_i)$, where
$\gamma_1,\dots , \gamma_r$ are such that $$\dotp{\Phi(x) - \sum_{i=1}^{r}
\gamma_i \Phi(z_i)}{\Phi(z_k)} = 0 \ \text{ for } k =1,\dots, r.$$

This implies the normal equations,
$$
K_s(x,z_k) = \sum_{i=1}^{r}\gamma_i K_s(z_i,z_k) \text{ for } k= 1, \dots, r,
$$
and implies that $\gamma_i$ can be explicitly computed as

\begin{align}
    \begin{pmatrix} \gamma_1 \\ \gamma_2 \\ \dots \\ \gamma_r \end{pmatrix} &= \begin{pmatrix} 
1 & K_s(z_1,z_2) & \dots & K_s(z_1,z_r)\\
K_s(z_1,z_2) & 1 & \dots & K_s(z_2,z_r)\\
\dots & \dots & \dots  & \dots \\
K_s(z_r,z_1) & K_s(z_r,z_2) & \dots & 1
\end{pmatrix}^{-1}\begin{pmatrix} K_s(x,z_1) \\ K_s(x,z_2) \\ \dots \\ K_s(x,z_r) \end{pmatrix}.
\end{align}

Since $\widehat{\Phi}(x)$ and $\Phi(x) - \widehat{\Phi}(x)$ are orthogonal, the squared norm of the residual is given by 
\begin{align*}
    \rho(x,s) &\defeq \|\widehat{\Phi}(x) - \Phi(x)\|_s^2 \\ 
    & = \|\Phi(x)\|^2_s - \|\widehat{\Phi}(x)\|^2_s \\ 
    &= 1 - \|\sum_{i=1}^r\gamma_i\Phi(z_i)\|^2_s.
\end{align*} 

This shows that $0 \leq \rho(x,s) \leq 1 $. Let $\psi(x,s) \defeq 1 - \rho(x,s)$. The closer $\psi(x,s)$
is to $1$, the lower the residual error is. From the preceding expression for $\gamma_i$ we can see that

\scalebox{0.8}{
\begin{minipage}{\linewidth}
        \begin{align}
            \label{tcrit:e:2}
            \psi(x,s) &=  
            \begin{pmatrix} K_s(x,z_1) \\ K_s(x,z_2) \\ \dots \\  K_s(x,z_r) \end{pmatrix}^T
            \begin{pmatrix} 
                1 & K_s(z_1,z_2) & \dots & K_s(z_1,z_r)\\
                K_s(z_1,z_2) & 1 & \dots & K_s(z_2,z_r)\\
                \dots & \dots & \dots  & \dots \\
                K_s(z_r,z_1) & K_s(z_r,z_2) & \dots & 1
            \end{pmatrix}^{-1}
            \begin{pmatrix} K_s(x,z_1) \\ K_s(x,z_2) \\ \dots \\ K_s(x,z_r) \end{pmatrix}.
        \end{align}
\end{minipage}
}

So, given training data $x_1,\dots,x_N$, if we define 
$$
g(s) \defeq \frac{1}{N}\sum_{i=1}^{N} \psi(x_i,s),
$$
then $g(s)$ is a measure of the loss of accuracy due to projection into $V_0$,
where $g(s)$ lies between $0$ and $1$ for each $s > 0.$ \emph{Higher} values of
$g(s)$ indicate \emph{lower} loss in precision due to projection into $V_0$.
In particular, if $g(s) = 1$,  there is absolutely no loss of precision. 
That is, 
$$\widehat{\Phi}(x_i) = \Phi(x_i) \text{ for }  1 \leq i \leq N.$$

\begin{figure}[h]
    \centering
    \subfloat[Validation Function] {
        \includegraphics[width=0.45\textwidth]{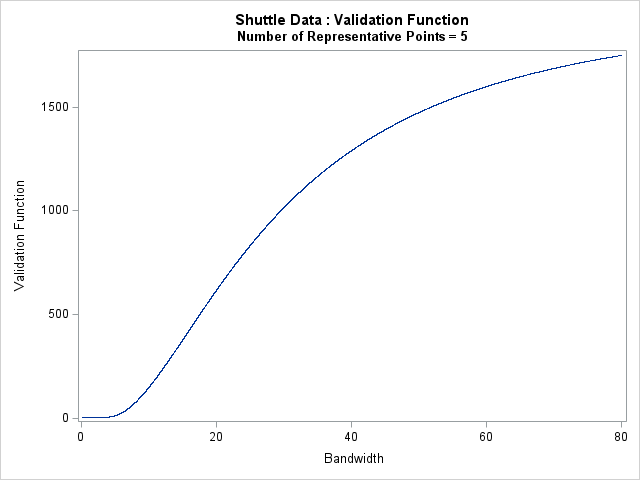}
    }
    \hfill
    \subfloat[Derivative of Validation Function]{
        \includegraphics[width=0.45\textwidth]{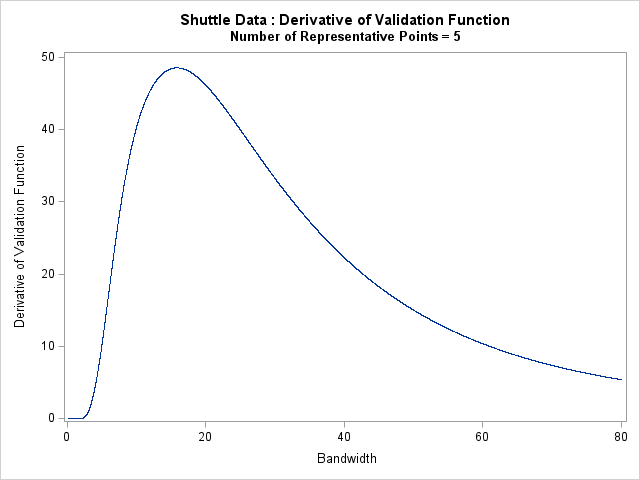}
    }
    \caption{Plot of a validation function and its derivative}
    \label{fig:1}
\end{figure}

It is empirically observed that typically $g(s)$ increases as $s$ increases (that is, $g'(s) > 0$). Moreover
$h(s) = g'(s)$ is unimodal: it initially increases until it reaches its maximum value, and then decreases. So $g(s)$
typically has a well-defined inflection point where $h(s)$ takes its maximum value. See Figure \ref{fig:1} for
the graph of a typical plot of $g()$ and $h()$.

The trace criterion method suggests using $s^* = \operatorname{argmax}_{s > 0} h(s)$ as the bandwidth that empirical
observations suggest coincides with the inflection point of $g$ in most cases.

Even though $h(s)$ is a scalar function, it is defined in terms of matrix operations and we can come up with an
explicit closed form for $h(s)$ by using usual rules of matrix differential calculus \citep{magnus1988matrix}.

Let $U(s)$ denote the $r \times r$ kernel matrix of representative points $(K_s(z_i,z_j))_{i,j=1}^r$. Its element-by-element derivative is given by
the $r \times r$ matrix,
$$U'(s) = \frac{1}{s^3}\left( \|z_i - z_j\|^2 K_s(z_i,z_j)\right)_{i,j=1}^r.$$

For $1 \leq i \leq N$, let $W_i(s) = (K_s(x_i,z_k))_{k=1}^r$ denote a $r \times 1$ column vector.
Then its element-by-element derivative is given by the $r \times 1$ column vector, $$ W_{i}'(s) = \frac{1}{s^3}
\left(\|x_i - z_k\|^2 K_s(x_i,z_k)\right)_{k=1}^r.$$

Then we have $$g(s) = \frac{\sum_{i=1}^N W_i^T(s) U^{-1}(s) W_i(s)}{N},$$ and hence 
\begin{equation*}
    h(s) = g'(s) = \frac{2 \sum_{i=1}^{N} W_i^T(s) U^{-1}(s) W_i'(s)}{N} - \frac{\sum_{i=1}^{N} W_i^T(s) U^{-1}(s) U'(s)
    U^{-1}(s) W_i(s)}{N}.
\end{equation*}

Let $B_i(s) \defeq U^{-1}(s) W_i(s)$. Then the preceding equation simplifies to
\begin{equation}
    h(s) = \frac{ 2 \sum_{i=1}^N B_i^T(s) W_i(s) }{N} - \frac{\sum_{i=1}^N B_i^T(s) U^{'}(s) B_i(s)}{N}.
    \label{eq:u}
\end{equation}

From \eqref{eq:u}, it is apparent that $h(s)$ is differentiable on $ 0 < s < \infty$ so we can use standard nonlinear
optimization methods such as Newton-Raphson to maximize it.

The cost of evaluating $h(s)$ is $\mathcal{O}(Nr^2)$, which is tractable because $r \ll N$.

We have to choose the number of representative points, $r$, for applying the
trace criterion method. We have observed that $r=5$ gives a good result for most
data sets, and we have used the trace criterion with $r=5$ throughout this paper
unless explicitly stated otherwise.

\section{Alternative Methods}
\label{sec:trealt}
Because SVDD is an unsupervised learning technique it is not possible
to use cross-validation to determine a suitable bandwidth value in
general. So it is desirable to use automatic, unsupervised bandwidth selection
techniques.

Such bandwidth selection methods for SVDD fall into two groups:
\begin{enumerate}
    \item Intrinsic methods:
        These methods suggest a bandwidth value solely on the basis of training
        data. Some intrinsic methods provide a closed-form expression for
        the bandwidth value, and some suggest a bandwidth value based on an
        iterative scheme or a bandwidth value obtained from solving an ancillary
        optimization problem.
    \item Selection methods that are based on edge-detection: 
        These methods use edge-detection techniques to determine the boundary of
        the data. The boundary points are used to generate pseudo-outliers that lie just
        beyond the boundary of the data. The pseudo-outliers along with the
        training data enables cross-validation to be used to determine a suitable
        bandwidth. In some methods, pseudo-inliers that mimic the training data
        are also generated and cross-validation based on the quality of fit on
        both the pseudo-outliers and the pseudo-inliers is used to determine a
        bandwidth.
\end{enumerate}

\subsection{Intrinsic Methods}
Common intrinsic methods for bandwidth selection for SVDD include: the
coefficient of variation method (CV) \cite{evangelista2007some}, the distance
to farthest neighbor method (DFN) \cite{xiao2014two}, the peak criterion method
\cite{2016arXiv160205257K,sergiy2017} and the modified mean criterion method (mean$_2$)
\cite{liao2018new} (see also \cite{ChaudhuriMean}). 
For the rest of this section, we assume that we are given data
$x_1,\dots,x_N \in \mathbb{R}^p$ and that $K(s) = (k_{ij}(s))_{i,j=1}^N =\left( \exp( - \dfrac{\| x_i - x_j
\|^2}{2s^2} ) \right)_{i,j=1}^N$ is the corresponding kernel matrix. 

\subsubsection{Coefficient of Variation method \cite{evangelista2007some}} 
The coefficient of variation (CV) method selects the bandwidth that maximizes
the following objective function:
\begin{equation}
f(s) = \frac{V(s)}{\kappa(s) + \epsilon}.
\end{equation}
Here, $V(s)$ is the variance of the non diagonal elements of the kernel matrix
and $\kappa(s)$ is their mean. $\epsilon$ is a pre-selected small quantity, which
is added to the denominator for numerical stability. The optimization can
be performed using gradient based methods. Because the computation of the
objective function entails an $\mathcal{O}(N^2)$ computational cost, this method
is computationally expensive for large data.
\subsubsection{Distance to Farthest Neighbor Method \cite{xiao2014two}}
The distance to farthest neighbor method (DFN) chooses the bandwidth that
maximizes the following objective function:
\begin{equation} 
    f(s) = \frac{1}{N} \sum_{i=1}^N \max_{j \neq i} k_{ij}(s) - \frac{1}{N} \sum_{i=1}^{N}
\min_{j} k_{ij}(s).
\end{equation}
The computation of the objective function
entails an $\mathcal{O}(N^2)$ computational cost, so this method is also
computationally expensive for large data.
\subsubsection{Peak Criterion Method\cite{2016arXiv160205257K}}
For a given bandwidth, say $s$, let $\theta(s)$ be the optimal value of the
objective function that defines the SVDD problem. The peak criterion method chooses
the bandwidth as the smallest zero of the second derivative of $\theta(s).$ The
second derivative is computed numerically by finite differences, by varying $s$
over a grid of values and solving the optimization problem that defines SVDD
for each value in the grid. The choice of the grid is very important because most
solvers run very slowly when supplied with a bandwidth value much smaller than
the scale of the values in the data. You also need to make sure the grid is
fine enough so that the second derivative can be computed to a reasonable level
of approximation. However making the grid fine means you have to solve the SVDD
optimization a large number of times which is computationally expensive.
\subsubsection{Modified mean Criterion \cite{liao2018new}}
The modified mean (mean$_2$) criterion method suggests a bandwidth which has a closed form in terms of the data.
The suggested bandwidth is given by 
$$
s = \sqrt{\frac{2N\sum_{i=1}^p \sigma^2_i}{(N-1)}}\sqrt{\frac{1}{\ln(\frac{N-1}{\delta^2})}}.
$$
Here, $\sigma_i^2$ is the variance of the $i^{\text{th}}$ variable and 
$$
\delta = -0.14818008\phi^4 + 0.284623624\phi^3 - 0.252853808\phi^2 + 0.159059498\phi - 0.001381145
$$
where $\phi = \frac{1}{\ln(N-1)}.$ 

The cost of the preceding computation is only $\mathcal{O}(Np).$
\subsection{Methods that are based on Edge Detection}
\subsubsection{Self-Adaptive Edge Detection Method \cite{wang2018hyperparameter}}
The current state of the art for a bandwidth selection method based on edge
detection is the self-adaptive data shifting method (SADS). This is a method that uses
K-nearest-neighbors (KNN) (with $5 \log_{10}(N)$
neighbors) to detect the points at lie at the boundary (edge) of the data along
with the outward normals at these points. This method uses the boundary points
to create pseudo-outliers that lie just beyond the data boundary. The next step
in this method is to estimate the direction of the maximum density for each
point in the data and use these directions to create a dataset of pseudo-inliers
that closely mimics the training data. Having generated the pseudo-outliers
and pseudo-inliers, the method uses cross-validation to select the bandwidth
that provides the best fit both on the pseudo-outliers and the pseudo-inliers.
The cross-validation is done by considering candidate bandwidth values over a
set that varies uniformly in the logarithmic scale. The authors of \cite{wang2018hyperparameter}
choose values of $\sigma = \sqrt{2}s$ that lie in the set $\{10^{-4},10^{-3},10^{-2},\dots,10^4\}$.
SADS is a ``fully automatic'' method, the choice of the number of neighbors and
the search grid is a part of the definition of SADS.

We have observed that in practice the default value of the number of neighbors
suggested by the authors of \cite{wang2018hyperparameter} is inappropriate for many datasets.
Also the default grid suggested in \cite{wang2018hyperparameter} is inadequate for many datasets. 
The logarithmic scale makes it likely that we skip the region near good bandwidth values for many datasets, for
example the method evaluates $\sigma$ values $10^3=1000$ and $10^4=10000$ in
succession. There will be many datasets for which a good value will lie between
these two values and neither of the suggested values are appropriate. We will
give an example of one such dataset later.

\section{Evaluating the Trace Criterion Method}
\label{sec:treval}
In this section we compare the trace criterion method with the five bandwidth selection
techniques mentioned in the previous section. The competing bandwidth selection
methods are tested on different types of datasets, both simulated and real, to
cover the kind of data that can arise in various situations. A 32.0 GB machine with 4 cores at 3.40 GHz was used for all analysis below.

\subsection{Choice of Data}
In the SVDD formulation, the objective function depends on the data through the
distance matrix. So it is natural to consider cases that lead to different types
of distance matrices. For example the distribution of pairwise distances of a
data where the observations lie in different clusters tends to be multimodal in
contrast to data where the observations are concentrated in one connected set
\cite{ChaudhuriMean}.

Another important factor than can impact the performance of different methods is
the dimension of the data. Methods that work well for two-dimensional data might
not work well for high-dimensional data, so it is important to compare the methods
on data of different dimensions. For two-dimensional data we can easily gauge
whether a bandwidth leads to a reasonable description by visual inspection.
However this cannot be done for higher dimensions. For high-dimensional data
we have two choices: 1) we can use labeled data where we have reasons to
believe that observations in the data corresponding to distinct labels lie in
geometrically separate regions of space, or, 2) we can use simulated data where
the ground truth is known.

With the preceding factors in mind we evaluate the trace criterion on the following four types of data:
\begin{itemize}
	\item Connected two-dimensional data: These are data that have two variables, and the observations 
        lie in a connected set, that is, there are no natural clusters in the data. 
	\item Disconnected two-dimensional data: These are data that have two variables and the observations 
        can be naturally divided into multiple clusters. 
	\item Higher-dimensional labeled real data: These are high-dimensional real data with multiple labels.
        We choose a subset of the data that corresponds to one label (usually the dominant label) and we see
        how well the SVDD fit that corresponds to a particular bandwidth separates our selected label from the 
        other labels.
    \item High-dimensional simulated data: We simulate high-dimensional data from various high-dimensional 
        geometric shapes. We consider the case where the observations are \emph{connected} and the case where
        the observations lie in multiple clusters. We simulate the data for different dimensions and study the
        impact of dimension on the different bandwidth selection methods.
\end{itemize}
\subsection{Comparison on two-dimensional data}
In this section we compare the performance of the different bandwidth selection
methods on various two-dimensional data. The two dimensional case is special
because we can easily make a judgement on the quality of description provided
by a bandwidth by visual inspection. We will compare the different methods on a
variety of \emph{connected} data, i.e., data without any obvious clusters, and
data with multiple clusters.

\subsubsection{Data Description}
This section compares the trace criterion with other bandwidth criteria
using two-dimensional connected data. Such data have no clusters. We use the
star-shaped data and banana-shaped data, and we train an SVDD model by using
the bandwidth value obtained by the peak, modified mean, trace criteria, CV,
DFN and SADS. To evaluate the results, we scored the bounding rectangle of the
data by dividing it into a 200 $\times$ 200 grid and plotted the inlier region.
With a good bandwidth value, the inlier region that is obtained from scoring
should match the geometry of the training data.

\subsubsection{Scoring Results}
Figures \ref{tab:2dcon1} and \ref{tab:2dcon2} display the results, along with
the bandwidth value. The inlier region is shaded darkly.
For banana-shaped data, the scoring results indicate that the bandwidth value
obtained using all methods except DFN provide good-quality data description. The
values of bandwidth $s$ obtained using CV, peak and trace are close. For
star-shaped data, CV and trace provide the best data description.

\begin{table}[h!]
    \captionof{figure}{Results for two-dimensional connected data (Banana-shaped data)}
    \label{tab:2dcon1}
    \begin{adjustbox}{width=\columnwidth,center}
        \begin{tabular}{cccc}
            \centering
            \multirow{2}{*}{\subfloat[Training Data]{\includegraphics[width=0.2\linewidth]{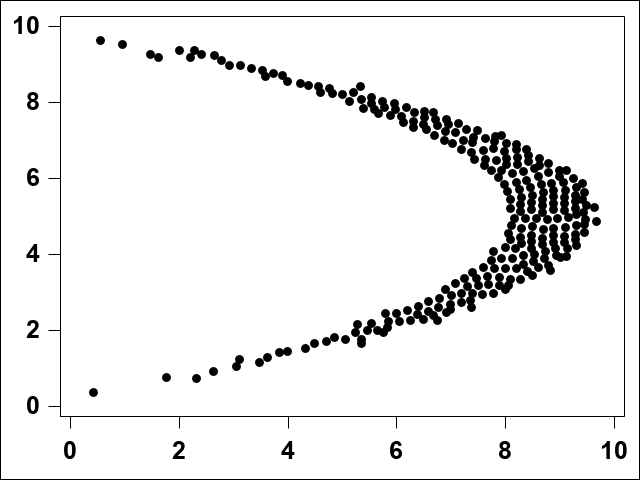}}} &
            \hfill
            \subfloat[CV $s=0.489$]{\includegraphics[width=0.2\linewidth]{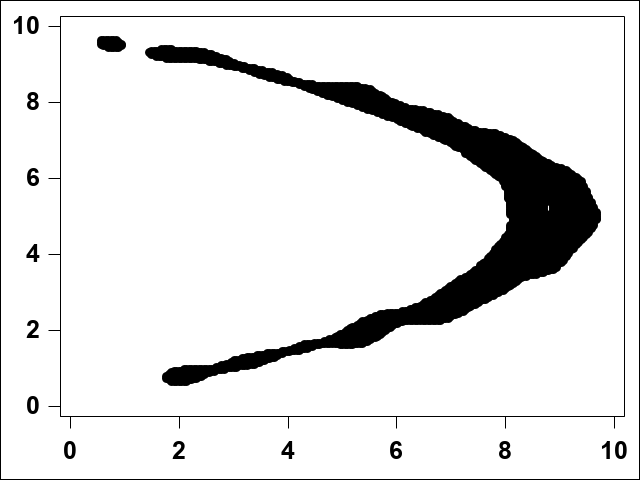}} &
            \hfill
            \subfloat[DFN $s=2.402$]{\includegraphics[width=0.2\linewidth]{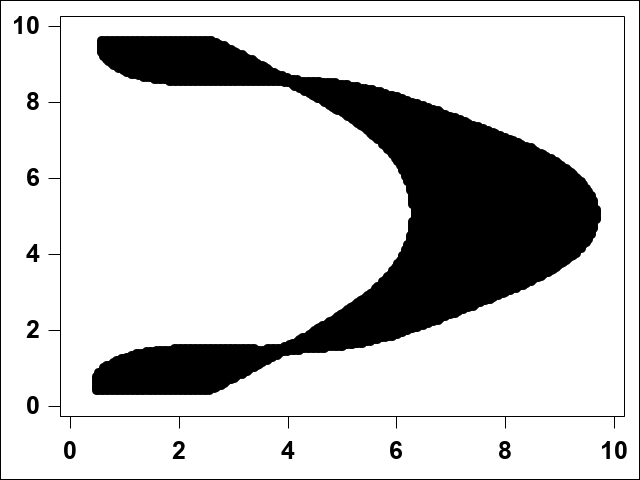}} &
            \hfill
            \subfloat[Peak $s=0.500$]{\includegraphics[width=0.2\linewidth]{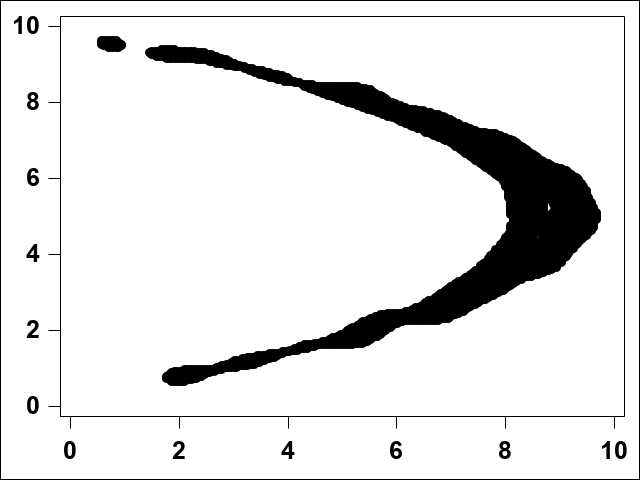}} \\
                                                                                                                                              &
                                                                                                                                              \hfill
            \subfloat[$\text{Mean}_2$ $s=1.145$]{\includegraphics[width=0.2\linewidth]{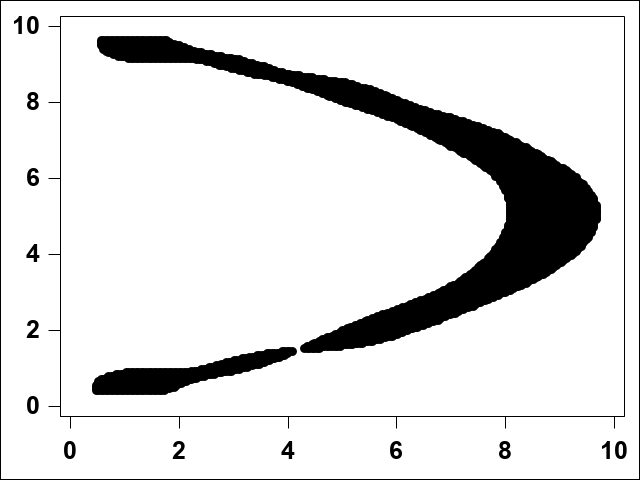}} &
            \hfill
            \subfloat[Trace $s=0.657$]{\includegraphics[width=0.2\linewidth]{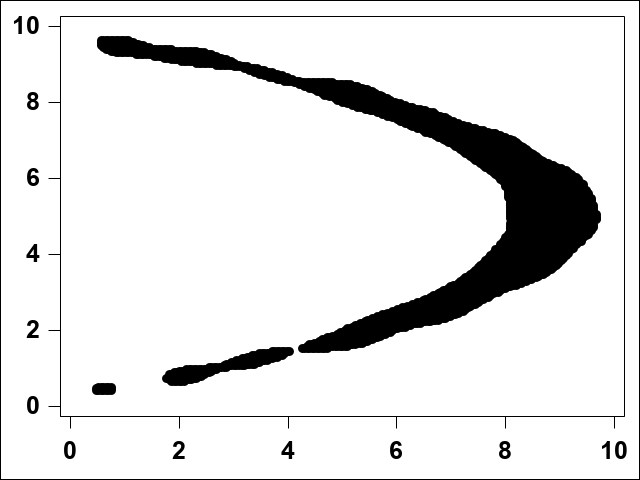}} &
            \hfill
            \subfloat[SADS $s=1.414$]{\includegraphics[width=0.2\linewidth]{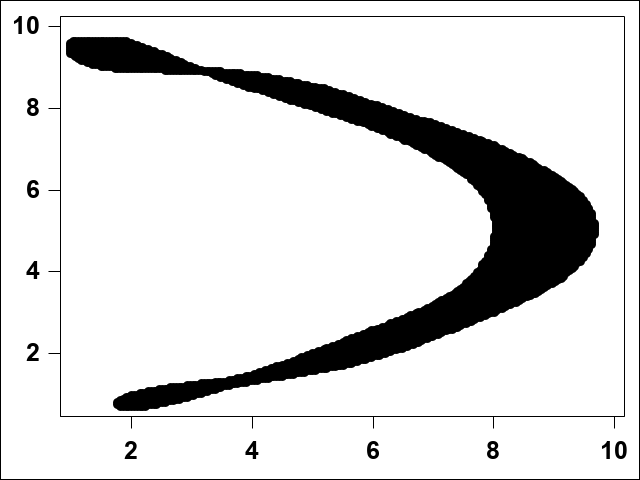}} \\
        \end{tabular}	
    \end{adjustbox}
\end{table}

\begin{table}
    \captionof{figure}{Results for Two-Dimensional Connected Data (Star-shaped Data)}
    \label{tab:2dcon2}	
    \begin{adjustbox}{width=\columnwidth,center}
    \begin{tabular}{cccc}
        \centering
        \multirow{2}{*}{\subfloat[Training Data]{
            \includegraphics[width=0.2\linewidth]{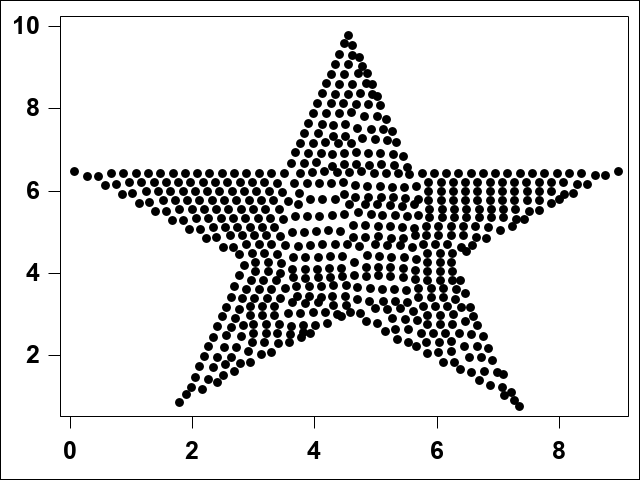}}
        } &
        \hfill
        \subfloat[CV $s=0.489$]{
            \includegraphics[width=0.2\linewidth]{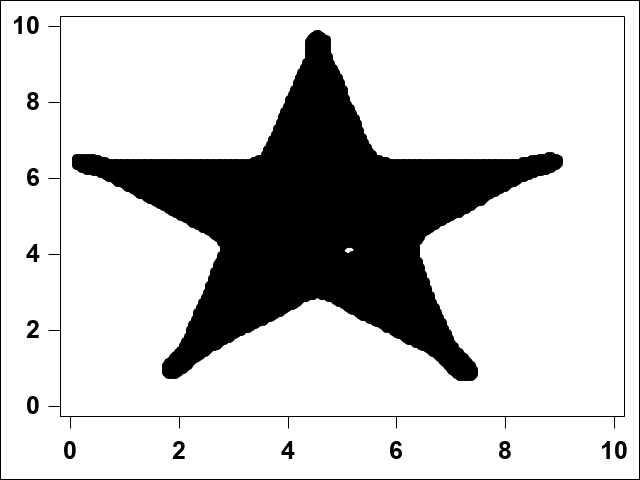}
        } &
        \hfill
        \subfloat[DFN $s=1.714$]{
            \includegraphics[width=0.2\linewidth]{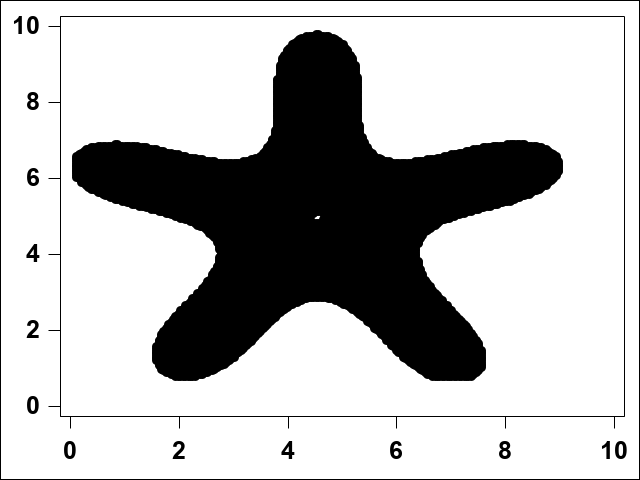}
        } &
        \hfill
        \subfloat[Peak $s=1.200$]{
            \includegraphics[width=0.2\linewidth]{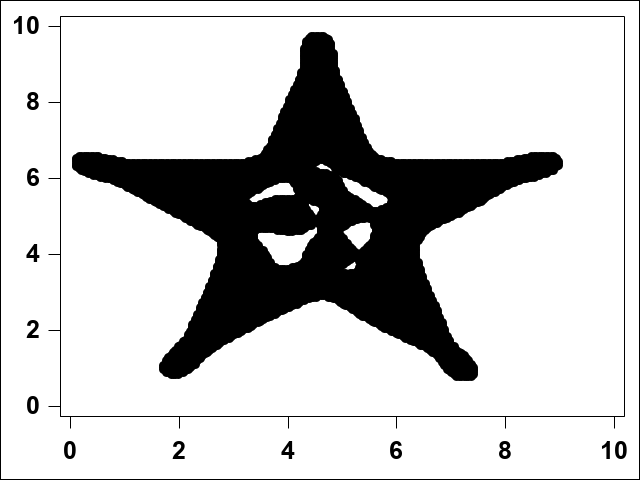}
        } \\
    &
    \hfill
    \subfloat[$\text{Mean}_2$ $s=0.936$]{
        \includegraphics[width=0.2\linewidth]{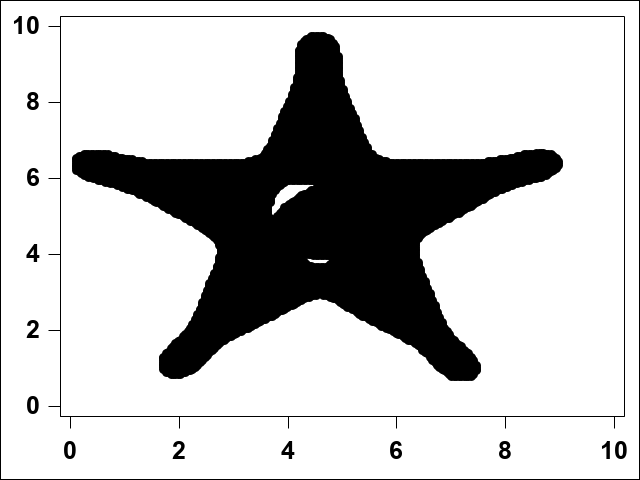} 
    } &
    \hfill
    \subfloat[Trace $s=0.846$]{
        \includegraphics[width=0.2\linewidth]{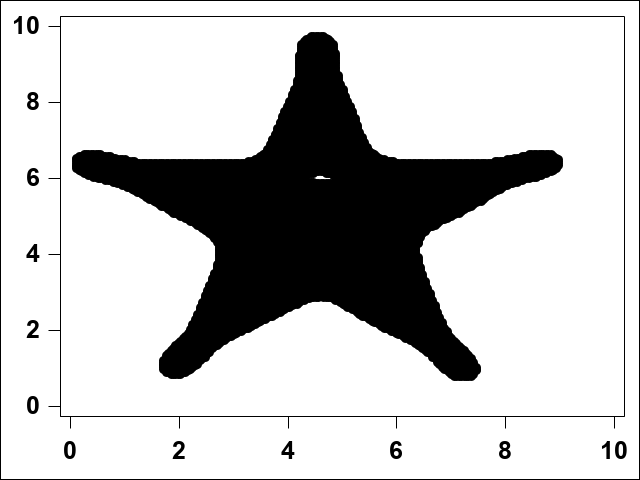}
    } &
    \hfill
    \subfloat[SADS $s=1.414$]{
        \includegraphics[width=0.2\linewidth]{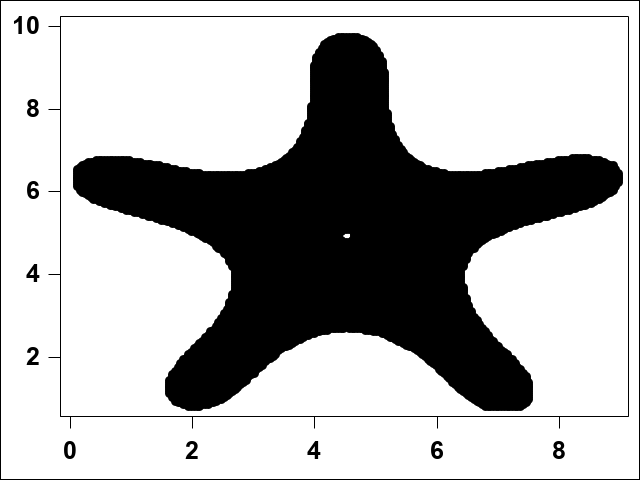}
    } \\
\end{tabular}	
\end{adjustbox}
\end{table}

\subsection{Comparison Using Two-Dimensional Disconnected Data}
\subsubsection{Data Description}
This section compares bandwidth selection methods using two-dimensional data
that lie across multiple clusters. We have observed that computing a good
bandwidth value for such data is more difficult than for connected data. Since
the data are two-dimensional, we can visually judge the quality of results. To
evaluate the results, we scored the bounding rectangle of the data by dividing
it into a 200 $\times$ 200 grid. With a good bandwidth value, the inlier region
obtained from scoring should match the geometry of the training data. We use the
following three data sets for evaluations:
\begin{itemize}
    \item The refrigerant data which consists of four clusters \cite{heck:2000} 
    \item A simulated ``two-donut and a circle'' data set, which consists of two
        donut-shaped clusters and one circular cluster
\end{itemize}

Figures \ref{tab:2ddiscon1} and \ref{tab:2ddiscon2} display the results, along
with the bandwidth value.
\subsubsection{Scoring Results}

\begin{table}[h]
    \captionof{figure}{Results for two-dimensional disconnected data (Refrigerant Data)}
    \label{tab:2ddiscon1}	
    \begin{adjustbox}{width=\columnwidth,center}
	\begin{tabular}{cccc}
		\centering
		\multirow{2}{*}{\subfloat[Training Data]{
 		\includegraphics[width=0.2\linewidth]{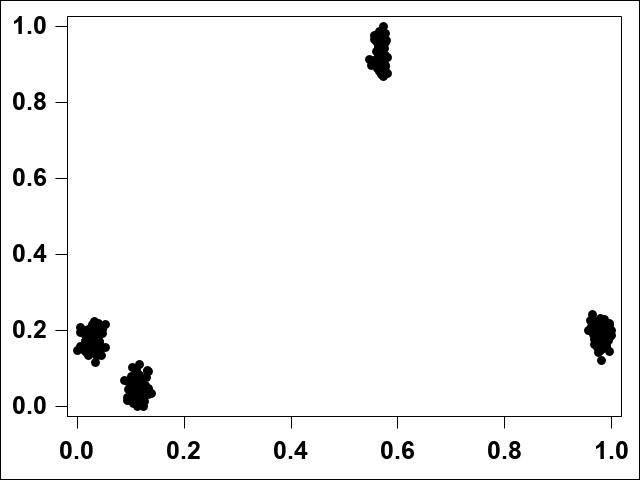}}
 	} &
    \hfill
 	\subfloat[CV $s=0.053$]{
 		\includegraphics[width=0.2\linewidth]{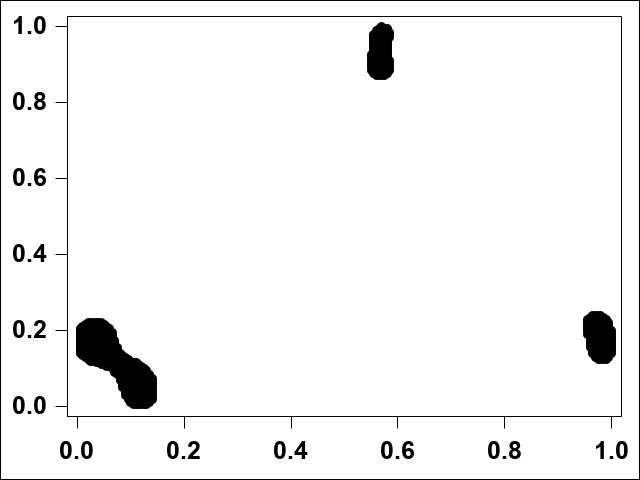}
 	} &
    \hfill
 	\subfloat[DFN $s=0.216$]{
 		\includegraphics[width=0.2\linewidth]{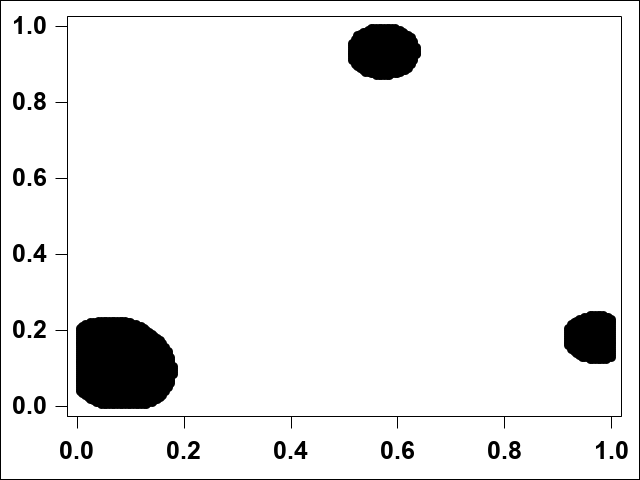}
 	} &
    \hfill
 	\subfloat[Peak $s=0.010$]{
 		\includegraphics[width=0.2\linewidth]{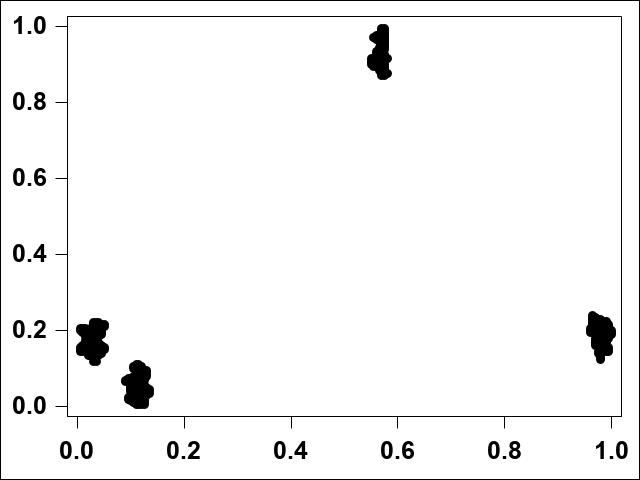}
 	} \\
 	&
    \hfill
 	\subfloat[$\text{Mean}_2$ $s=0.197$]{
 		\includegraphics[width=0.2\linewidth]{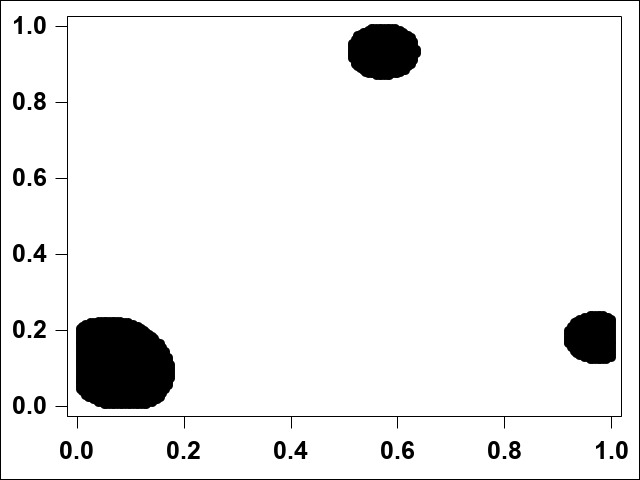} 
 	} &
    \hfill
 	\subfloat[Trace $s=0.011$]{
 		\includegraphics[width=0.2\linewidth]{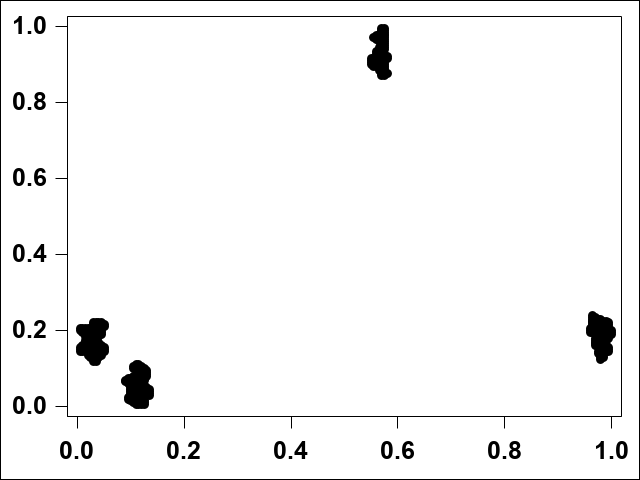}
 	} &
    \hfill
 	\subfloat[SADS $s=0.014$]{
 		\includegraphics[width=0.2\linewidth]{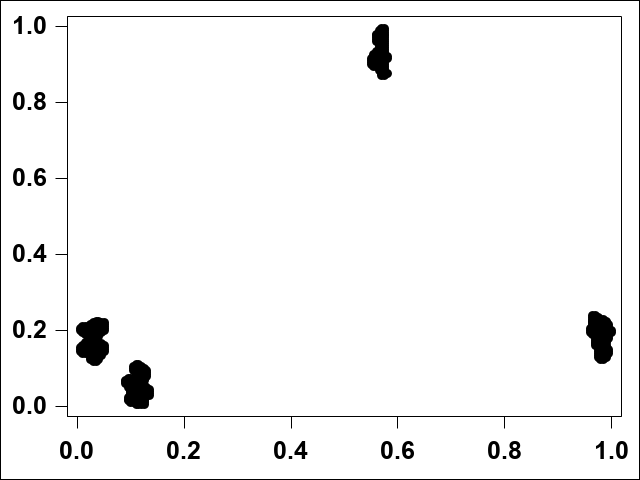}
 	} \\
 	\end{tabular}
\end{adjustbox}
 \end{table}
 
 \begin{table}[h]
     \captionof{figure}{Results for two-dimensional disconnected data (Two Donuts and a circle data). Note that the CV bandwidth is very small, almost zero, thus, none of the observations in the scoring grid are labeled as inliers.}
     \label{tab:2ddiscon2}	
     \begin{adjustbox}{width=\columnwidth,center}
         \begin{tabular}{cccc}
             \centering
             \multirow{2}{*}{\subfloat[Training Data]{
                 \includegraphics[width=0.2\linewidth]{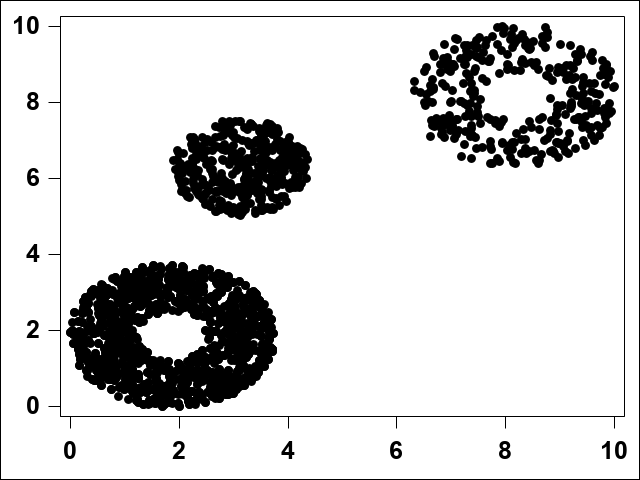}}
             } & 
             \hfill
             \subfloat[CV $s = 0.0006$]{
                 \includegraphics[width=0.2\linewidth]{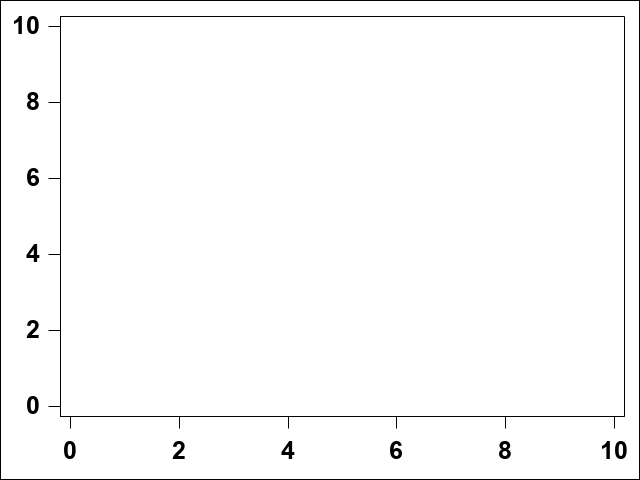}
             } &
             \hfill
             \subfloat[DFN $s=1.799$]{
                 \includegraphics[width=0.2\linewidth]{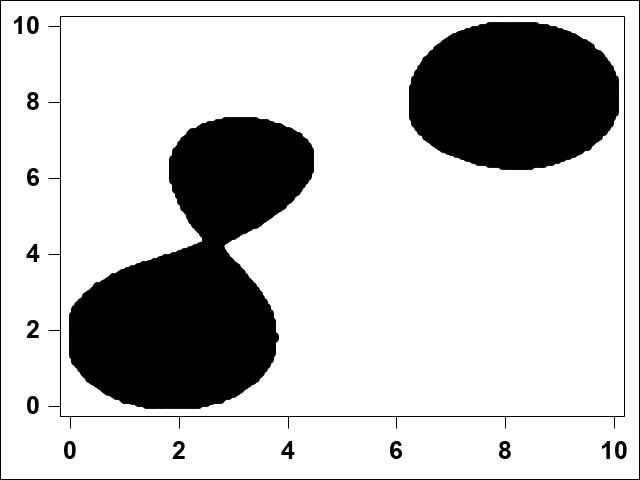}
             } &
             \hfill
             \subfloat[Peak $s=0.800$]{
                 \includegraphics[width=0.2\linewidth]{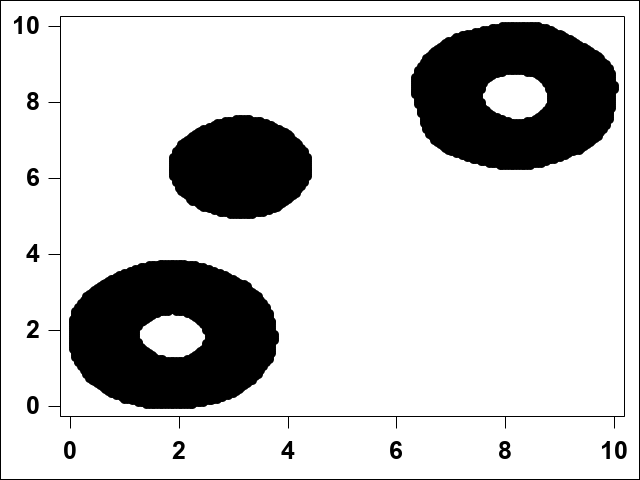}
             } \\
    &
    \hfill
    \subfloat[$\text{Mean}_2$ $s=1.249$]{
        \includegraphics[width=0.2\linewidth]{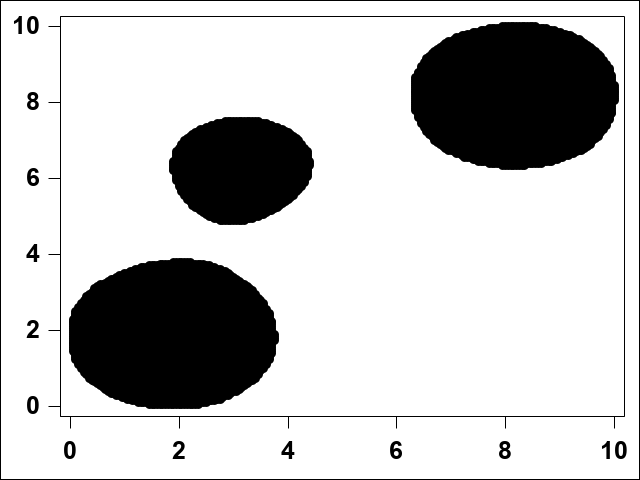} 
     } &
     \hfill
     \subfloat[Trace $s=0.794$]{
         \includegraphics[width=0.2\linewidth]{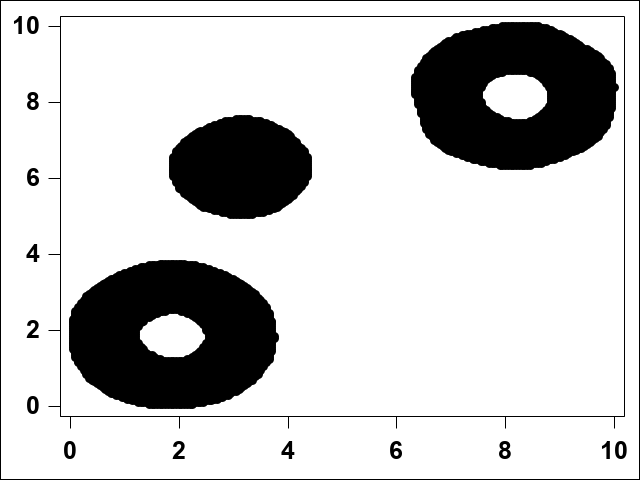}
     } &
     \hfill
     \subfloat[SADS $s=0.141$]{
         \includegraphics[width=0.2\linewidth]{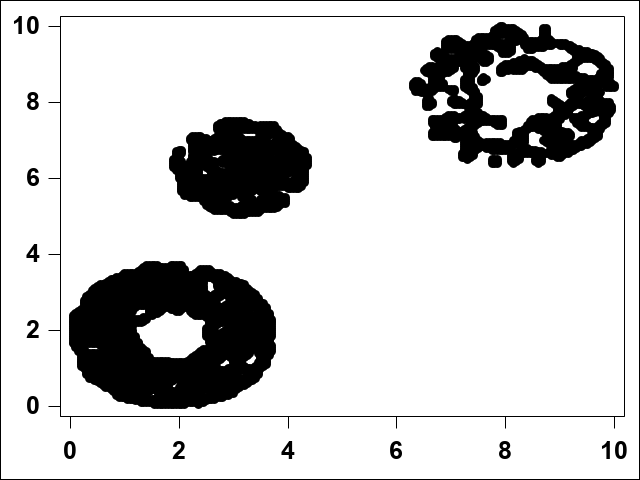}
     } \\
 \end{tabular}
\end{adjustbox}
 \end{table}

For the refrigerant data, the data description that is obtained using peak, trace
and SADS is able to separate out all four clusters, whereas the description
obtained using the modified mean criterion, CV and DFN merges the two clusters
that lie close to each other.

For the ``two-donut and a circle'' data set, the peak and trace methods choose an
appropriate bandwidth and deliver the best visual results. The DFN and modified
mean methods both choose a large bandwidth, while CV and SADS select a small
bandwidth.

\subsection{Comparison Using High-Dimensional Data}
Comparing the different criteria for high-dimensional data is
much more difficult than comparing them for two-dimensional
data. In two-dimensional data, the quality of the result can be
easily judged by looking at the plot of the scoring results.
But this is not possible for high-dimensional data. For the
purpose of evaluation, we selected labeled high-dimensional
data that have a dominant class. We used SVDD on a subset
of the dominant class to obtain a description of it, and then we scored the rest of the data to evaluate
the criteria. We expected the points in the scoring data set that
correspond to the dominant class to be classified as inliers
and all other points to be classified as outliers. Because the
data are labeled, we could also use cross validation to determine
the bandwidth that best describes the dominant class in the
sense of maximizing a measure of fit, such as the $F_1$ score.
So in this section we compare the bandwidths that are suggested by
the different unsupervised criteria with the bandwidth that is obtained
through cross validation for various benchmark data sets. The
results are summarized in Table \ref{tab:hd} . The benchmark data
sets used for the analysis are described in sections \ref{subsec:shuttle}.\\

\begin{table}[h]
    \centering
    \renewcommand{\arraystretch}{1.1}
    \caption{Results for high-dimensional Data}
    \label{tab:hd}	
    \begin{adjustbox}{width=0.8\columnwidth,center}
        \begin{tabular}{cccccccccccccc}\toprule
            \multicolumn{4}{c}{{Method}} & \phantom{abc} & \multicolumn{4}{c}{Shuttle} & \phantom{abc} &
            \multicolumn{4}{c}{\makecell{Tennessee \\ Eastman}}\\ \cmidrule{1-4} \cmidrule{6-9} \cmidrule{11-14}
            \multicolumn{4}{c}{\phantom{abc}} & \phantom{abc} & \multicolumn{2}{c}{$F_1$} & \multicolumn{2}{c}{$s$} & \phantom{abc} &
            \multicolumn{2}{c}{$F_1$} & \multicolumn{2}{c}{$s$} \\  \cmidrule{6-9} \cmidrule{11-14}
            \multicolumn{4}{c}{Cross-Validation} & \phantom{abc} & \multicolumn{2}{c}{0.96} &
            \multicolumn{2}{r}{17.0} & \phantom{abc} & \multicolumn{2}{c}{0.18} & \multicolumn{2}{r}{12.25} \\
            \multicolumn{4}{c}{CV} & \phantom{abc} & \multicolumn{2}{c}{0.93} &
            \multicolumn{2}{r}{9.2} & \phantom{abc} & \multicolumn{2}{c}{0.15} & \multicolumn{2}{r}{7.3} \\
            \multicolumn{4}{c}{DFN} & \phantom{abc} & \multicolumn{2}{c}{0.91} &
            \multicolumn{2}{r}{225.0} & \phantom{abc} & \multicolumn{2}{c}{0.17} & \multicolumn{2}{r}{24.1} \\
            \multicolumn{4}{c}{Peak} & \phantom{abc} & \multicolumn{2}{c}{0.81} &
            \multicolumn{2}{r}{5.0} & \phantom{abc} & \multicolumn{2}{c}{0.15} & \multicolumn{2}{r}{7.0} \\
            \multicolumn{4}{c}{Modified Mean} & \phantom{abc} & \multicolumn{2}{c}{0.96} &
            \multicolumn{2}{r}{17.2} & \phantom{abc} & \multicolumn{2}{c}{0.18} & \multicolumn{2}{r}{11.4} \\
            \multicolumn{4}{c}{SADS} & \phantom{abc} & \multicolumn{2}{c}{0.95} &
            \multicolumn{2}{r}{14.1} & \phantom{abc} & \multicolumn{2}{c}{0.18} & \multicolumn{2}{r}{14.1} \\
            \multicolumn{4}{c}{Trace} & \phantom{abc} & \multicolumn{2}{c}{0.96} &
            \multicolumn{2}{r}{13.1} & \phantom{abc} & \multicolumn{2}{c}{0.18} & \multicolumn{2}{r}{11.2} \\
            \bottomrule
        \end{tabular}
    \end{adjustbox}
\end{table}

\subsubsection{Data Description}

\paragraph{Shuttle Data} \label{subsec:shuttle}
This data set consists of measurements made on a shuttle.
The data set contains nine numeric attributes and one classification
attribute. Of 58,000 total observations, 80\% 
belong to class 1. A random sample of 2,000
observations belonging to class 1 was selected for training, and
the remaining 56,000 observations were used for scoring. This
data set is from the UC Irvine Machine Learning Repository \cite{Dua:2019}.

\paragraph{Tennessee Eastman Data} \label{subsec:te}
This data set was generated using MATLAB simulation
code, which provides a model of an industrial chemical
process. The data were generated for normal operations of
the process and 20 faulty processes. Each observation
consists of 41 variables, out of which 22 were measured
continuously every 6 seconds on average and the remaining
19 were sampled at a specified interval of every 0.1 or 0.25
hours. From the simulated data, we created an analysis data
set that uses the normal operations data of the first 90 minutes
and data that correspond to faults 1 through 20. A data set
that contains observations of normal operations was used for
training. Scoring was performed to determine whether the
model could accurately classify an observation as belonging
to normal operation of the process. The MATLAB simulation
code is available at \cite{Ricker:2002}.

\subsubsection{Scoring Results}
The results outlined in table \ref{tab:hd} show that
the bandwidths by the proposed trace, modified mean and SADS methods each have a $F_1$ measure 
value that is almost identical to that given by the bandwidth selected using cross validation.

\subsection{Comparison Using Simulated Data}

In this section, we present results of a simulation study that we conducted
to compare performance of different methods.
The simulations were performed to generate training data with a known
geometry. The inlier region corresponding to a good bandwidth selection method should exactly match the
known true geometry of underlying data. And we can use the discrepency from the true geometry to rank different
bandwidth selection methods.

The data dimensions were varied between 2 and 40. We conducted five
simulation studies. Table \ref{tab:sims} provides details of these studies.

\begin{table}[h]
    \centering
    \caption{Simulation study description}
    \label{tab:sims}
    \renewcommand{\arraystretch}{1.3}
    \begin{adjustbox}{width=0.9\columnwidth,center}
        \begin{tabular}{p{4cm}@{\hskip 0.5cm}p{2cm}@{\hskip 0.5cm}p{2cm}@{\hskip 0.5cm}p{5cm}}\toprule
            \centering Simulation & \centering Data Dimension & \centering Number of data sets & Description \\ \hline
            Two-dimensional polygons & 2 & 500 & Polygons with varying number of vertices and length of sides\\ \hline
            Hypercubes & 5 to 40 in increments of 5 & 200  & Single hypercube used for training\\ \hline
            Hyperspheres & 5 to 40 in increments of 5 & 200 & Single hypersphere used for training\\ \hline
            High-dimensional disconnected data with multiple hyperspheres & 5 to 40 in increments of 5 & 160 & Each training
            data set contains multiple disjoint hyperspheres, either 5 or 10.\\ \hline
            High-dimensional disconnected data with multiple hypercubes & 5 to 40 in increments of 5  & 160 & Each training
            data set contains multiple disjoint hypercubes, either 5 or 10.\\ \bottomrule
        \end{tabular}
    \end{adjustbox}
\end{table}

\subsubsection{Evaluation Using Two-Dimensional Polygons}
In this section, we measure the performance of the trace criterion when it is applied to randomly generated polygons. Given
the number of vertices, $k$, we generate the vertices of a polygon in the anticlockwise sense as $r_1
\exp i \theta_{(1)}, \dots, r_k \exp i \theta_{(k)}.$ Here $\theta_{(1)} = 0$ and $\theta_{(i)}$ for $i = 2,\dots,n$ are the order statistics of an i.i.d sample
that is uniformly drawn from $(0,2\pi).$ The $r_i$ are uniformly drawn from an interval $[\text{r}_{\text{min}},\text{r}_{\text{max}}].$ 

\begin{figure}
    \centering
	\caption{Random polygons}\label{fig:randpoly}
    \subfloat[Number of Vertices = 5]{
        \includegraphics[width=0.3\textwidth]{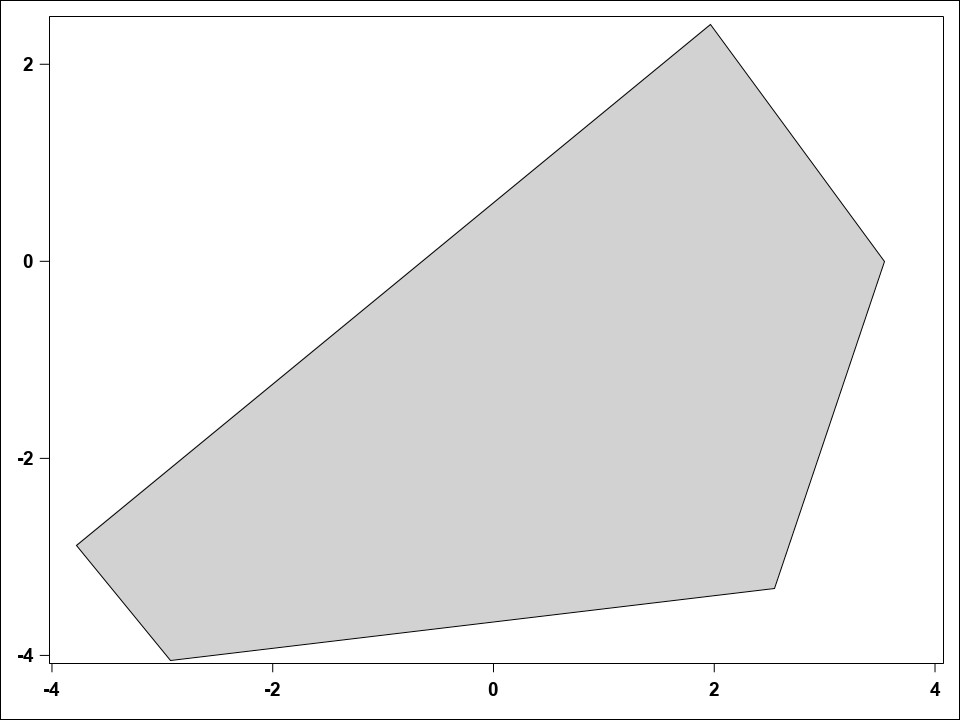}
    }
	\subfloat[Number of Vertices = 15]{
		\includegraphics[width=0.3\textwidth]{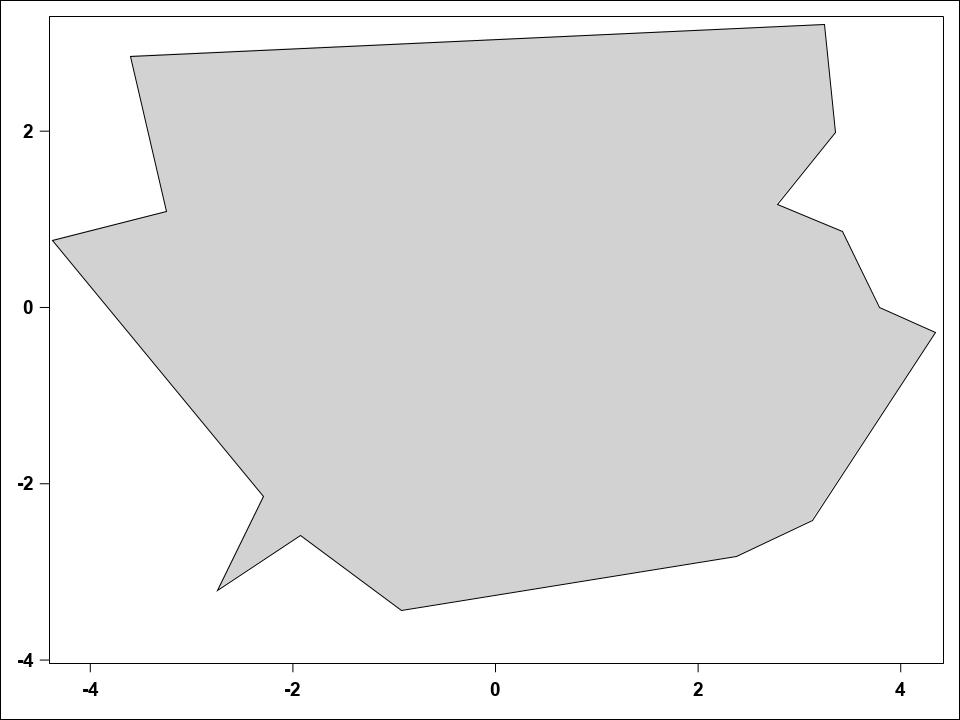}
	}
    \subfloat[Number of Vertices = 30]{
        \includegraphics[width=0.3\textwidth]{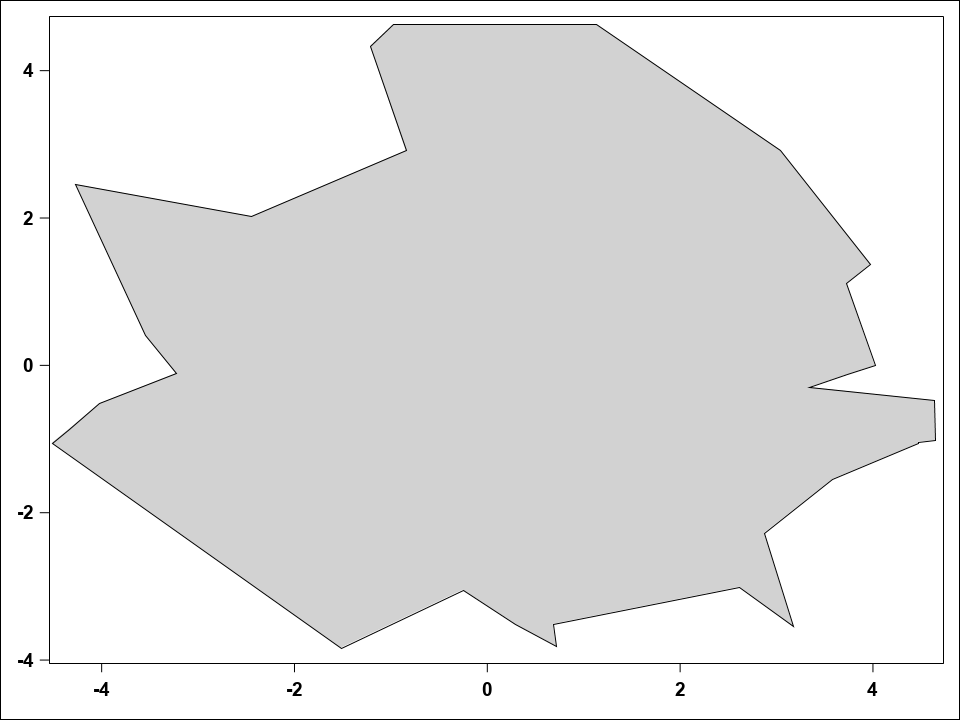}
    }
\end{figure}

For this simulation, we chose $\text{r}_{\text{min}}=3$ and
$\text{r}_{\text{max}}=5$ and varied the number of vertices from $5$ to $30$. We generated $20$ random polygons for each
vertex size. Having determined a polygon, we randomly sampled $1200$ points uniformly from the interior of the
polygon and used trace criterion and this sample to determine a bandwidth value. Figure \ref{fig:randpoly} shows three random polygons.

The box and whiskers plot in Figure \ref{fig:Tracepolygons}a summarizes the simulation
study results. The X-axis shows the
number of vertices of the polygon and Y-axis shows the $F_1$ score. The
bottom and the top of the box shows the first and the third quartile values.
The ends of the whiskers represent the minimum and the maximum values of the
$F_1$ score. The plot shows that the $F_1$ score is greater
than 0.9 across all values of number of vertices for all methods except CV. 
Because the complexity of
the polygon increases as the number of vertices increases, we observed that
the $F_1$ score decreased slightly. The fact that
$F_1$ score is always greater than 0.9 provides necessary evidence that
the trace criterion performs well. Figure \ref{fig:Tracepolygons}b summarizes the time performance of each method. For all number of vertices, modified mean and trace out-perform CV, DFN, peak and SADS, showing that trace is competitive in both accuracy and performance.

\begin{table}[h!]
    \captionof{figure}{Simulation results for random polygons}
    \label{fig:Tracepolygons}
    \begin{adjustbox}{width=\columnwidth,center}
        \begin{tabular}{cc}
            \subfloat[Accuracy using random polygons]{\includegraphics[width=0.4\textwidth]{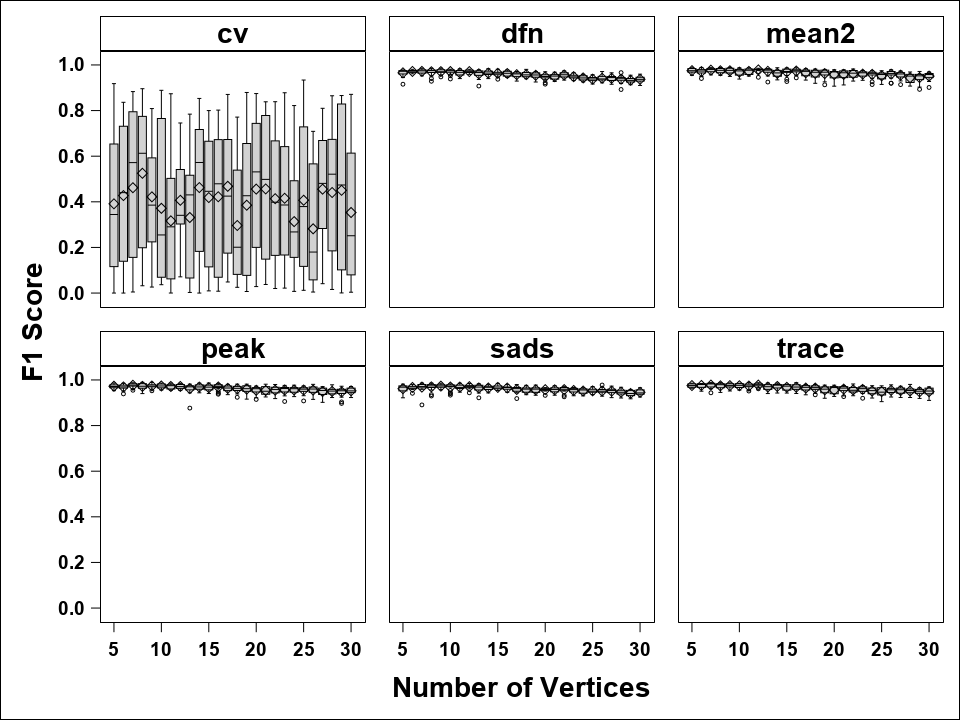}} & \subfloat[Timing using random polygons]{\includegraphics[width=0.4\textwidth]{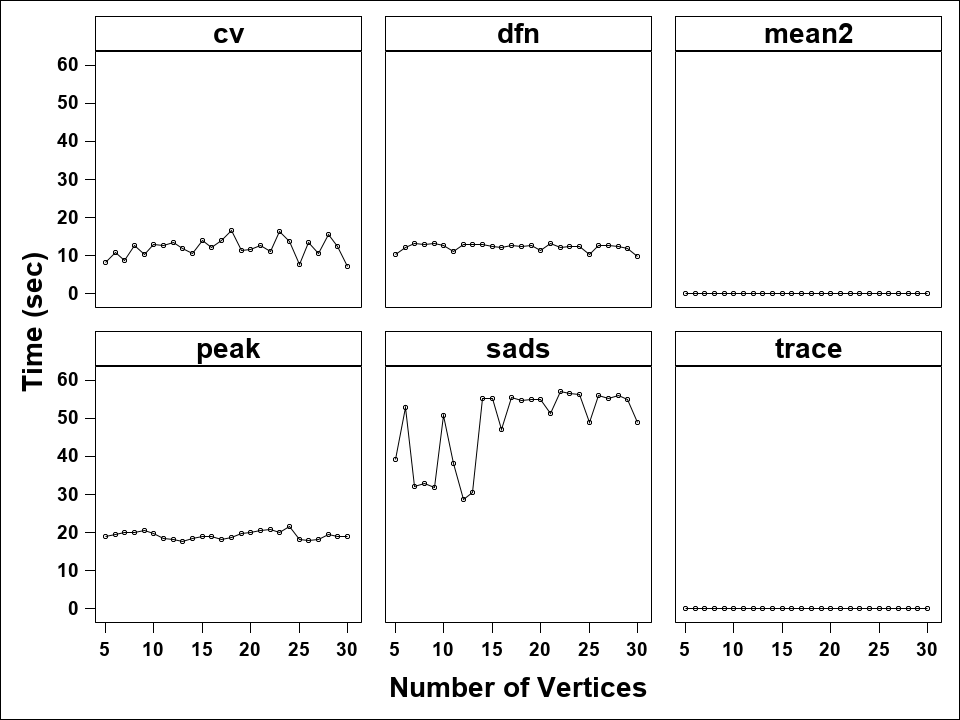}}\\
        \end{tabular}
    \end{adjustbox}
\end{table}

\subsubsection{Evaluation Using Hyperspheres}
\label{sec:spheres}
In this section, we evaluate the trace criterion by using spherical data of
varying dimensions. The observations in such spherical data (or hypersphere)
are uniformly distributed. We use scoring to evaluate the quality of the
data description that is obtained using the trace criterion bandwidth value.
The scoring data set consists of 50\% \emph{inlier} observations, which are
uniformly distributed inside the training sphere and 50\% \emph{outlier}
observations, which are uniformly distributed outside the sphere. The points
outside the sphere lie in a narrow annular ring, just outside the sphere.
Figure \ref{fig:hyperspheres} illustrates two variables in the training and
scoring data. The rationale behind creating such scoring data set is that
if the bandwidth value is good, then the data description that is obtained
using such a value should be able to discriminate between observations that
are inside and observations that are just outside the sphere. We varied the
hypershpere dimension from 5 to 40 in increments of 5. For each dimension, 25
sets of training and scoring data sets were simulated. We computed the $F_1$
measure for each data set to determine the quality of data description. The width
of the annular region was half the radius of the sphere.

Figure \ref{fig:sphere_boxwhisker}a shows a box-and-whiskers plot of the $F_1$ measure
for various values of data dimension. The $F_1$ measure decreases as the number
of variables (the data dimension) increases from 5 to 40. For CV, DFN, peak and trace methods,
the $F_1$ measure is
consistently above 0.9 for all simulated data sets across different dimensions.
While for the modified mean method, 
the $F_1$ measure dropped rapidly with the increase in the hypersphere
dimension. The SADS method fails to produce a bandwidth value for dimensions greater than 25. The best observed bandwidth for these high dimensional spheres is $0.6<s<1$. Note that this range of values is not in the SADS grid $[10^{-4}\sqrt{2}, 10^{-3}\sqrt{2},\cdots, 10^{4}\sqrt{2}]$. This is the greatest disadvantage of SADS; if the best bandwidth value does not lie near the specified grid points, only pseudo-outliers will be generated resulting in a failure to produce a bandwidth.
Figure \ref{fig:sphere_boxwhisker}b shows the amount of time taken to compute each bandwidth value. Note that the trace and modified mean bandwidths are computed in under 1 second, while CV, DFN, peak, and SADS are computed in over 20 seconds. 
These observations confirm that a bandwidth value that is obtained using
the trace criterion provides competitive-quality data description with
competitive-time performance.

\begin{table}[h!]
    \captionof{figure}{Evaluation using hyperspheres}\label{fig:hyperspheres}
    \centering
    \begin{adjustbox}{width=0.8\columnwidth,center}
        \begin{tabular}{cc}
            \subfloat[Training Data, \#obs=5,000]{
                \includegraphics[width=0.3\columnwidth]{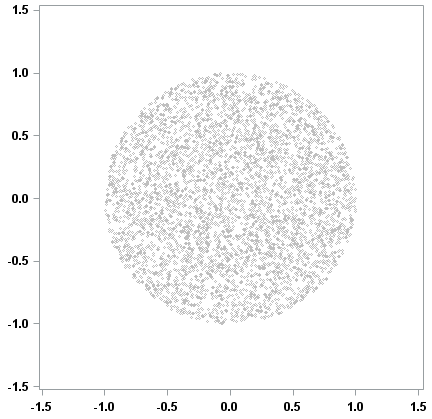}
            }
    &
    \subfloat[Scoring Data, \#obs=10,000]{
        \includegraphics[width=0.3\columnwidth]{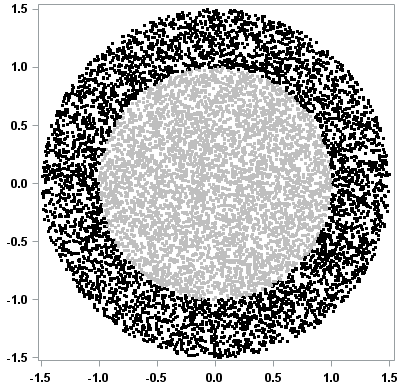}
    }
\end{tabular}
\end{adjustbox}
\end{table}

\begin{table}
    \captionof{figure}{Simulation results for hyperspheres. SADS fails to obtain a bandwidth value for all dimensions greater than 25}
    \label{fig:sphere_boxwhisker}
    \begin{adjustbox}{width=\columnwidth,center}
        \begin{tabular}{cc}
            \subfloat[Evaluation using hyperspheres.]{\includegraphics[width=0.4\textwidth]{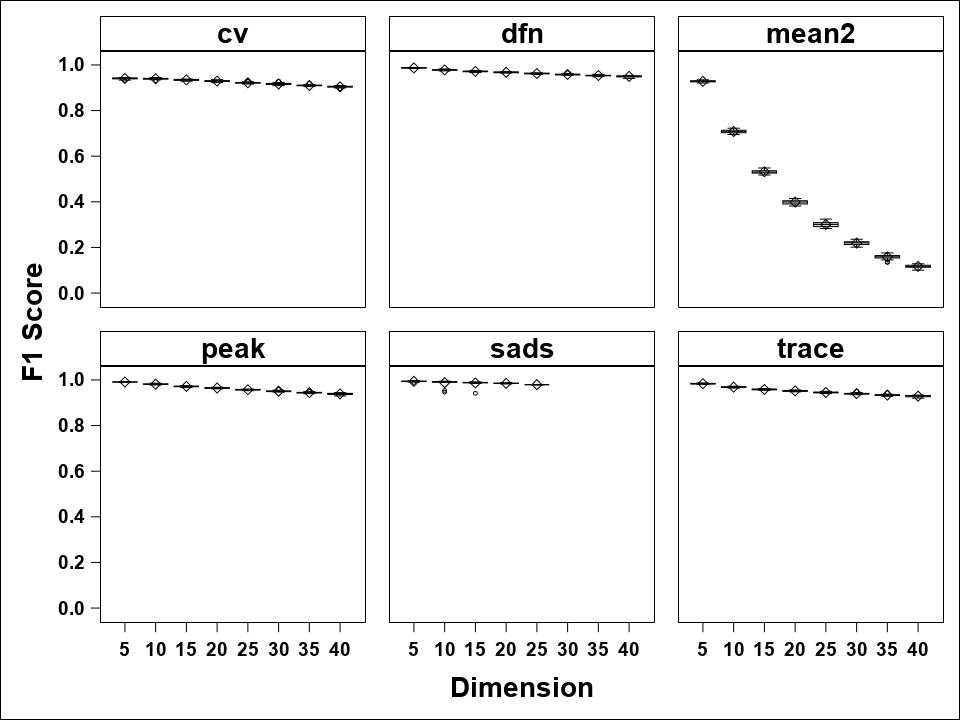}} & \subfloat[Timing using hyperspheres]{\includegraphics[width=0.4\textwidth]{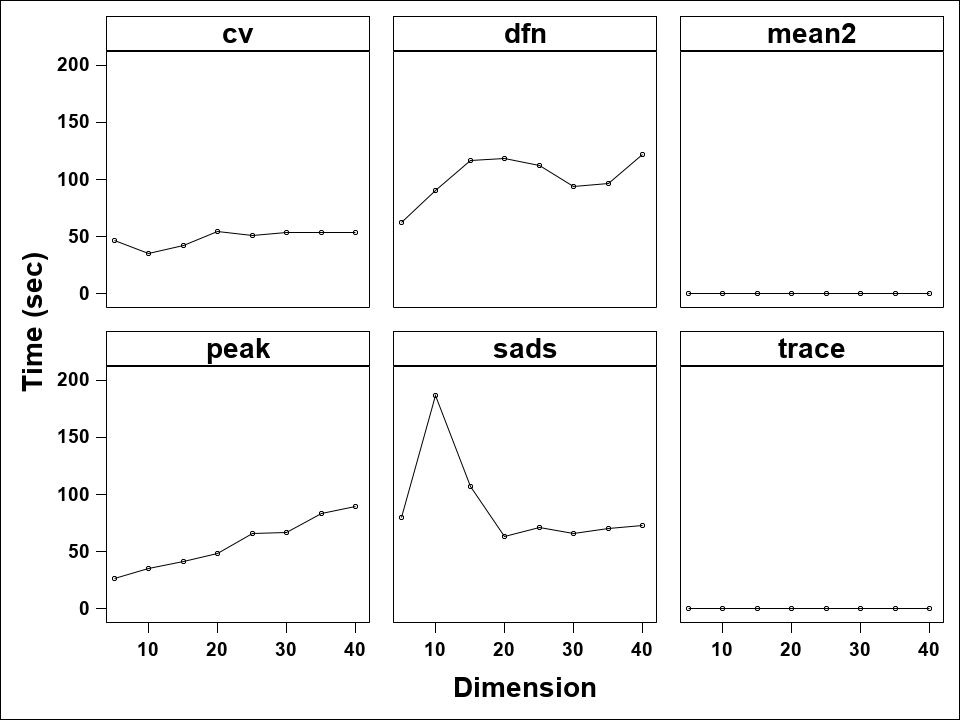}}\\
        \end{tabular}
    \end{adjustbox}
\end{table}

\subsubsection{Evaluation Using Hypercubes}
\label{sec:cubes}
In this section, we evaluate the trace criterion using cube-shaped data
of varying dimensions. The observations in such cubic data (or hypercube)
are uniformly distributed. We used scoring to evaluate the quality of data
description that was obtained using the trace criterion bandwidth value.
The scoring data set consists of 50\% \emph{inlier} observations, which
are uniformly distributed inside the training cube and 50\% \emph{outlier}
observations which are uniformly distributed outside the cube. The points
outside the cube lie in a narrow frame, just outside the cube. Figure
\ref{fig:hypercubes} illustrates the two variables in the training and
scoring data. The rationale behind creating such a scoring data set is that
if the bandwidth value is good, then the data description that is obtained
using such a value should be able to discriminate between observations that
are inside and observations that are just outside the cube. We varied the
hypercube dimension from 5 to 40 in increments of 5. For each dimension, 25
sets of training and scoring data sets were simulated. The width of the frame was 0.25
times the length of the cube. We computed the $F_1$
measure for each data set to determine the quality of data description. Figure
\ref{fig:cube_boxwhisker}a shows a box-and-whiskers plot of the $F_1$ measure
for various values of data dimension. We observed 
a similar pattern as in the hypersphere case.  The $F_1$ measure decreases as the number
of variables (the data dimension) increases from 5 to 40. For CV, DFN, peak and trace method,
the $F_1$ measure is
consistently above 0.7 for all simulated data sets across different dimensions.
While for modified mean method, 
the $F_1$ measure value dropped rapidly with the increase in the hypercube
dimension. The SADS method fails to converge to a bandwidth for all dimensions greater than 25. This observation confirms
that bandwidth values obtained using the trace criterion provide a much
better-quality data description than the modified mean
and SADS criteria provide. 
Moreover, figure \ref{fig:cube_boxwhisker}b shows that the modified mean and trace methods compute the bandwidth in less time than CV, DFN, peak, and SADS, thus illustrating that the trace criteria is competitive in both accuracy and performance.

\begin{table}[h!]
    \centering
    \captionof{figure}{Evaluation using hypercubes}\label{fig:hypercubes}
    \begin{adjustbox}{width=0.8\columnwidth,center}
        \begin{tabular}{cc}
            \subfloat[Training Data, \#obs=5,000]{
                \includegraphics[width=0.4\linewidth]{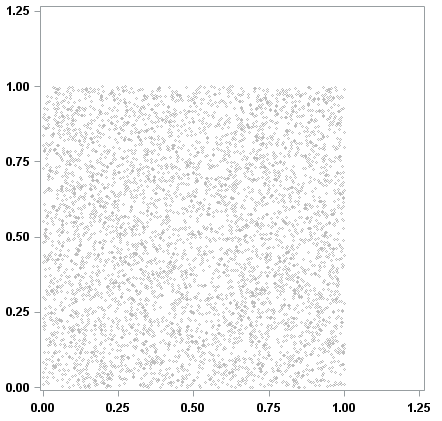}
            }
        &
        \subfloat[Scoring Data, \#obs=10,000]{
            \includegraphics[width=0.4\linewidth]{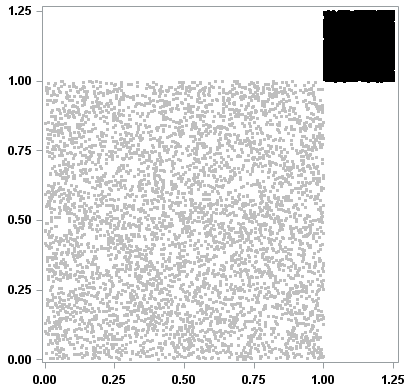}
        }
    \end{tabular}
\end{adjustbox}
\end{table}

\begin{table}[h!]
    \centering	
    \captionof{figure}{Simulation results using hypercubes. Peak fails to obtain a bandwidth value for $50\%$ of cubes of dimensions 20 or greater. SADS fails for $100\%$ of cubes of dimensions 30 or greater.}\label{fig:cube_boxwhisker}
    \begin{adjustbox}{width=\columnwidth,center}
        \begin{tabular}{cc}
            \subfloat[Evaluation using hypercubes]{
                \includegraphics[width=0.4\textwidth]{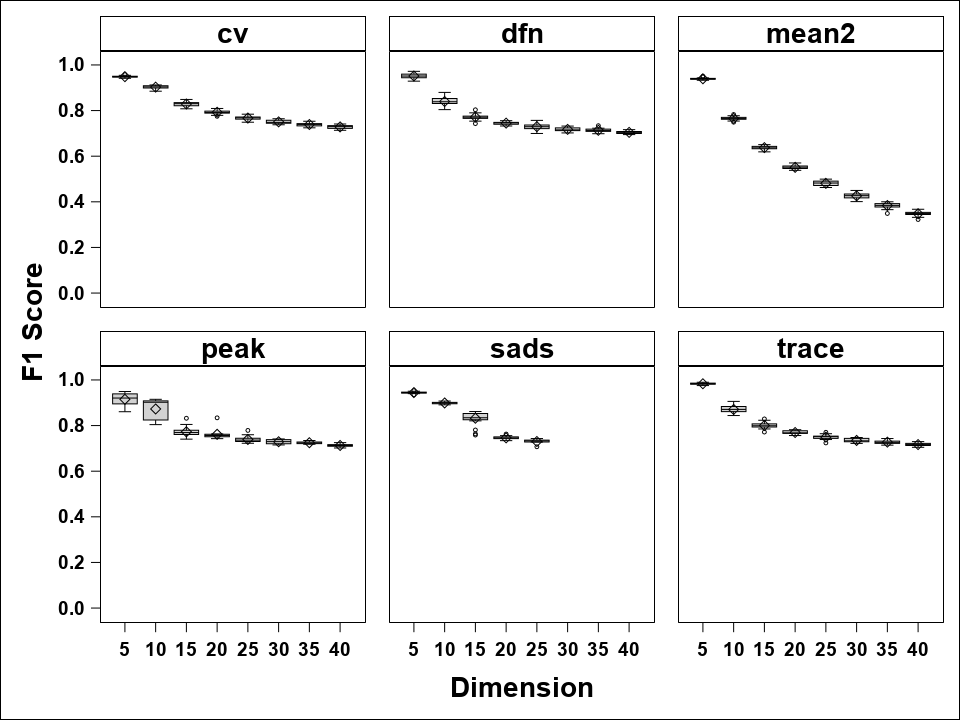}
            }
    &
    \subfloat[Timing using hypercubes]{
        \includegraphics[width=0.4\textwidth]{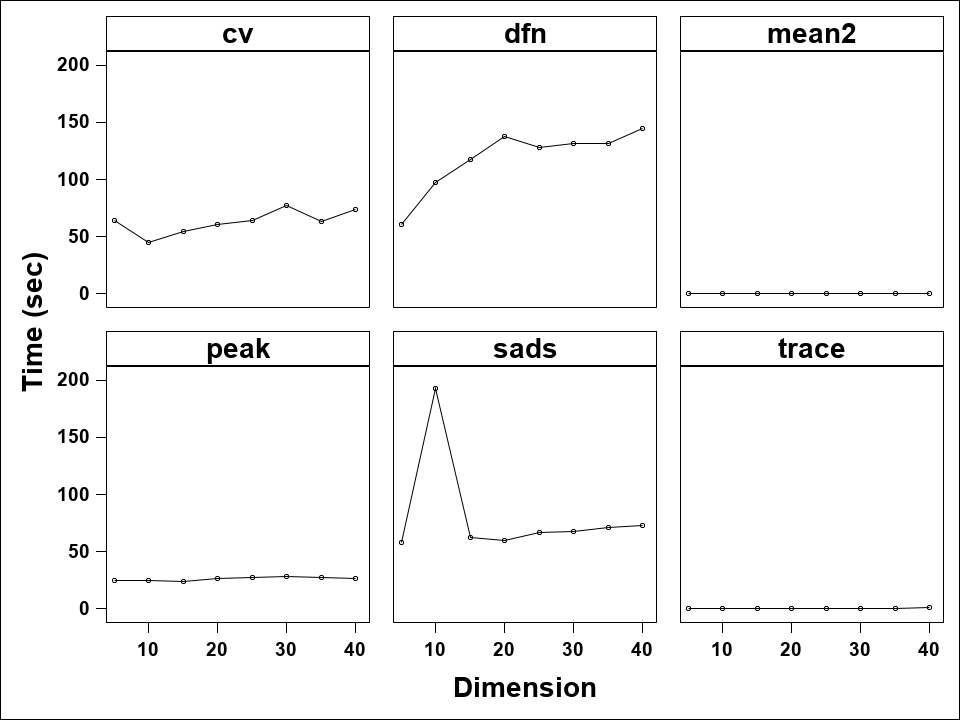}
    }

\end{tabular}
\end{adjustbox}
\end{table}

\subsubsection{Evaluation using Multiple Hyperspheres}
In this section, we use high-dimensional disconnected data to evaluate the performance of the trace criteria. The training data consist of two or more disjoint hyperspheres used earlier in section \ref{sec:spheres}. 

We evaluated the bandwidth criteria by using data sets that contain multiple numbers of spheres. The number of spheres in a data set was either 5 or 10, and the data dimension ranged from 5 to 40. For each combination of data dimension and number of spheres, we used different seed values to generate 10 different sets of training and scoring data sets. Figure \ref{fig:spheresmm} illustrates sample training and scoring data sets that have five spheres and use two dimensional data. 

\begin{table}
    \centering
    \captionof{figure}{Evaluation using multiple hyperspheres}\label{fig:spheresmm}
    \begin{adjustbox}{width=0.8\columnwidth,center}
    \begin{tabular}{cc}
        \subfloat[Training Data]{	
            \includegraphics[width=0.4\columnwidth]{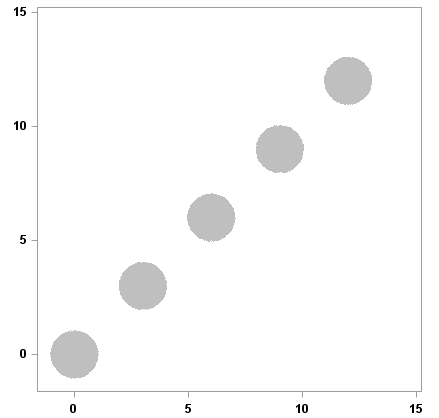}
        }
        &
        \subfloat[Scoring Data]{	
            \includegraphics[width=0.4\columnwidth]{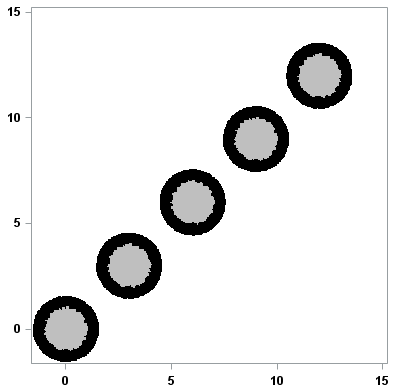}
        }
    \end{tabular}
\end{adjustbox}
\end{table}

For each simulation run, we computed the $F_1$ measure. Figure
\ref{fig:multiSpheres5} and Figure \ref{fig:multiSpheres10} provide the
box-and-whiskers plot of the $F_1$ measure for simulations results that were
obtained using 5 and 10 hyperspheres respectively. The CV, peak, and trace
criteria perform well on this simulation setting, giving F1 scores consistently
higher than 0.8. Additionally, CV frequently fails to converge to a bandwidth
value, and SADS does not provide any bandwidth values for spheres with dimension greater than 20. Trace outperforms both CV and peak in time performance.

\begin{table}
    \centering	
    \captionof{figure}{Simulation results using 5 hyperspheres.  CV fails to converge to a bandwidth value for $60\%$ of hyperspheres of dimensions 5 or greater. SADS fails to converge for $80\%$ of hyperspheres of dimension 20 and for $100\%$ of hyperspheres of dimensions 25 or greater.} \label{fig:multiSpheres5}
    \begin{adjustbox}{width=\columnwidth,center}
        \begin{tabular}{cc}
            \subfloat[Evaluation using 5 hyperspheres.]{ 
                \includegraphics[width=0.4\textwidth]{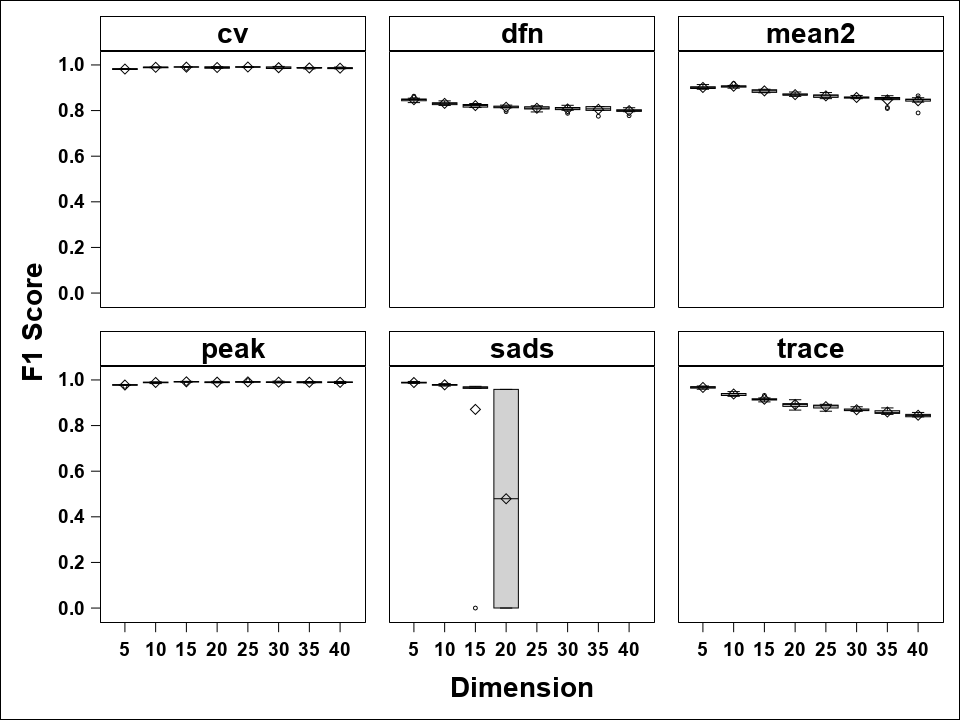}
            }
    &
    \subfloat[Timing using 5 hyperspheres]{
        \includegraphics[width=0.4\textwidth]{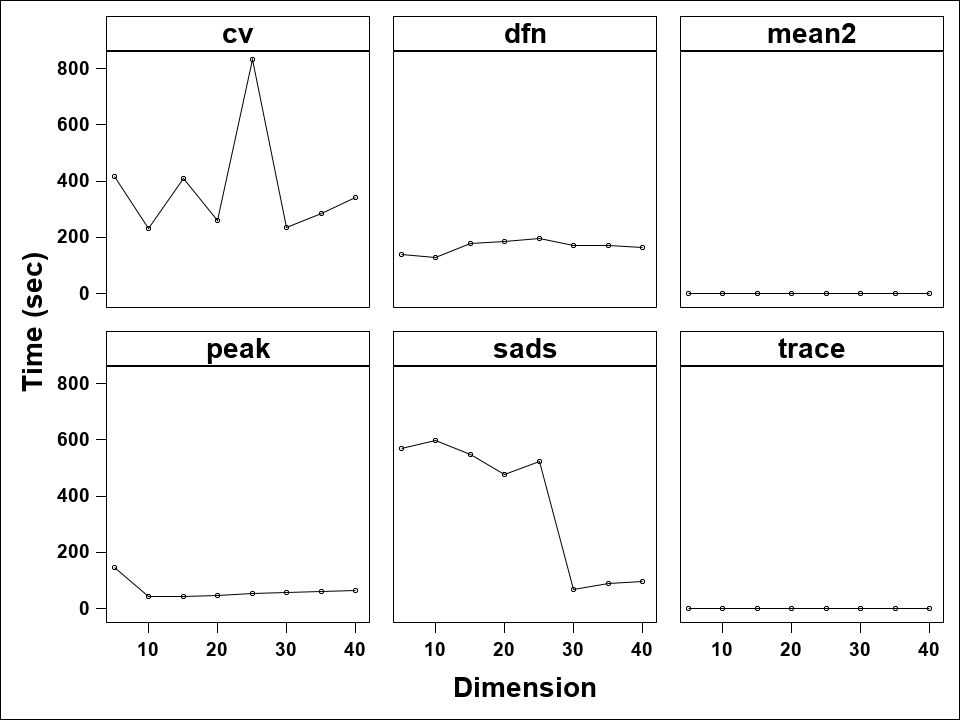}
    }
\end{tabular}
\end{adjustbox}
\end{table}

\begin{table}
	\centering	
    \captionof{figure}{Simulation results using 10 hyperspheres. CV fails to converge to a bandwidth value for $70\%$ of hyperspheres of dimensions 5 or greater. SADS fails to converge for $100\%$ of hyperspheres of dimensions 20 or greater.}\label{fig:multiSpheres10}
    \begin{adjustbox}{width=\columnwidth,center}
    \begin{tabular}{cc}
	\subfloat[Evaluation using 10 hyperspheres.]{ 
		\includegraphics[width=0.4\textwidth]{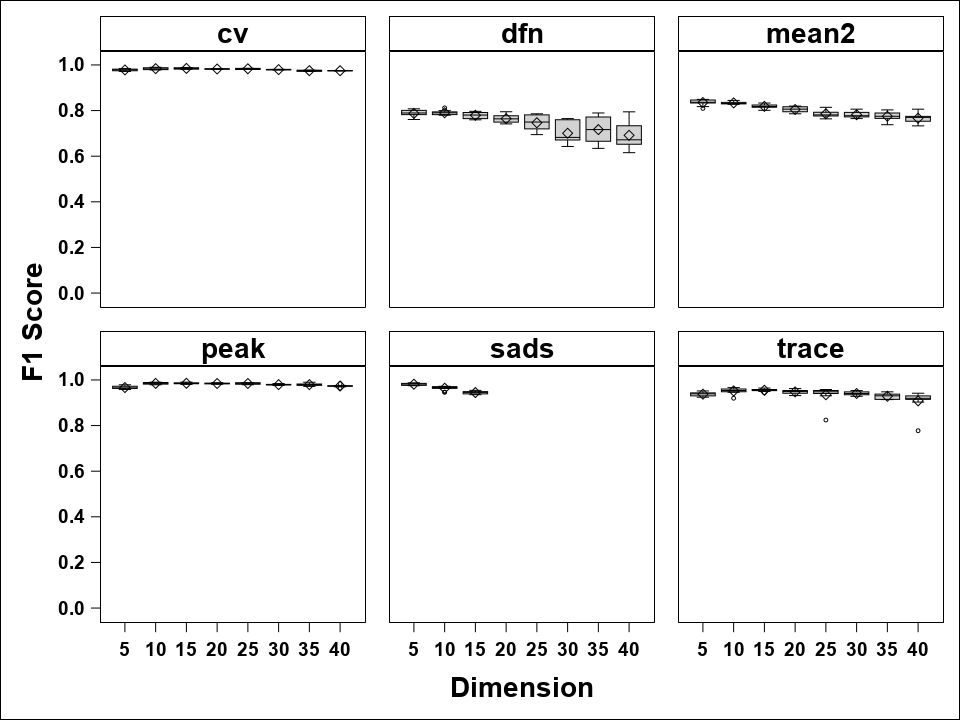}
	}
    &
	\subfloat[Timing using 10 hyperspheres]{
		\includegraphics[width=0.4\textwidth]{graphs/SimulationGraphs/hdmultspheres_incDim_10spheres_time2}
	}
\end{tabular}
\end{adjustbox}
\end{table}

\subsubsection{Evaluation Using Multiple Hypercubes}
We evaluated the bandwidth criteria by using data sets that contain multiple numbers of cubes. The number of cubes in a data set was either 5 or 10, and the data dimension was varied between 5 and 40 in increments of 5. For each combination of data dimension and number of cubes, we generated 10 different sets of training and scoring data sets by different seed value. Figure \ref{fig:cubesmm} illustrates sample training and scoring data set that have five cubes and use two-dimensional data. 

\begin{table}[h!]
	\centering
	\caption{Evaluation using multiple hypercubes}\label{fig:cubesmm}
    \begin{adjustbox}{width=0.8\columnwidth,center}
    \begin{tabular}{cc}
    \subfloat[Training Data]{
		\includegraphics[width=0.4\columnwidth]{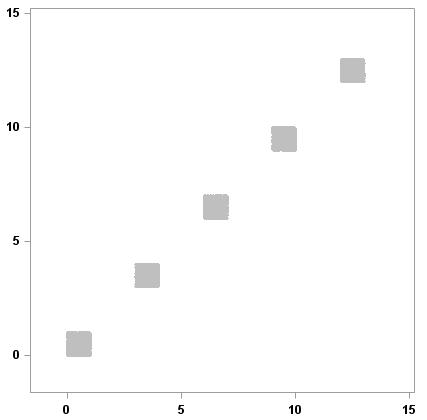} 
    } &
    \subfloat[Scoring Data]{
	    \includegraphics[width=0.4\columnwidth]{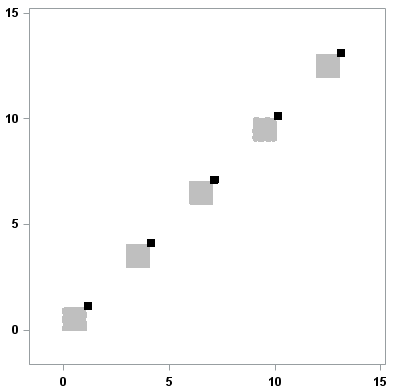}
    }
\end{tabular}
\end{adjustbox}
\end{table}

\begin{table}
    \centering	
    \captionof{figure}{Simulation results using 5 hypercubes. CV fails to converge to a bandwidth value for $60\%$ of hypercubes of dimension 5 or greater. SADS fails to converge for $100\%$ of hypercubes of dimension 25 or greater.}\label{fig:multiCubes5}
    \begin{adjustbox}{width=\columnwidth,center}
    \begin{tabular}{cc}
        \subfloat[Evaluation using 5 hypercubes.]{ 
            \includegraphics[width=0.5\textwidth]{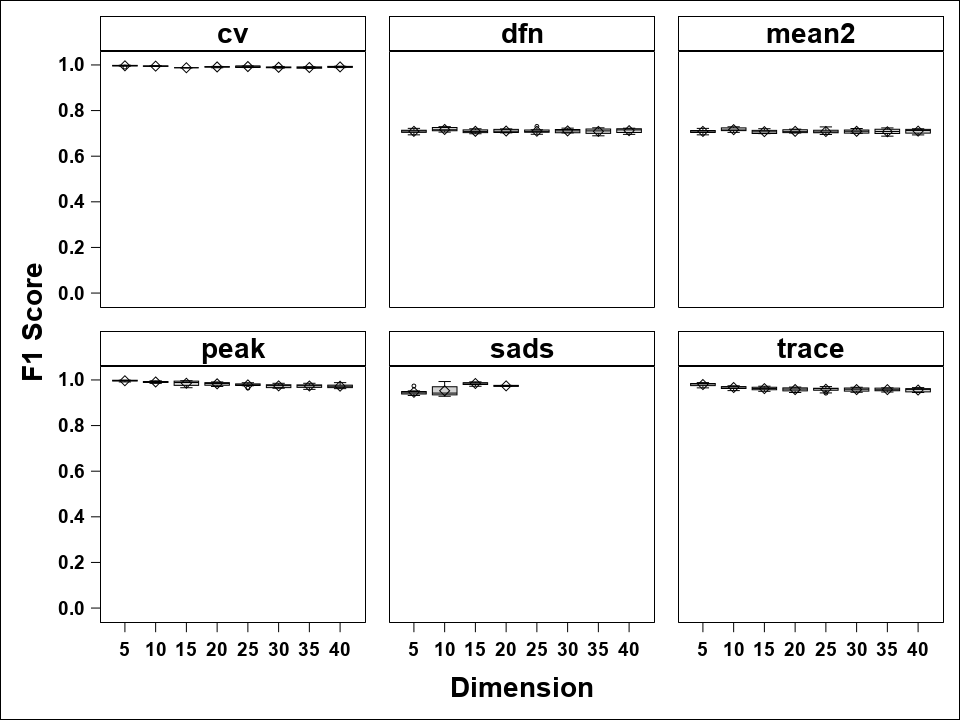}
        }
        \subfloat[Timing using 5 hypercubes]{ 
            \includegraphics[width=0.5\textwidth]{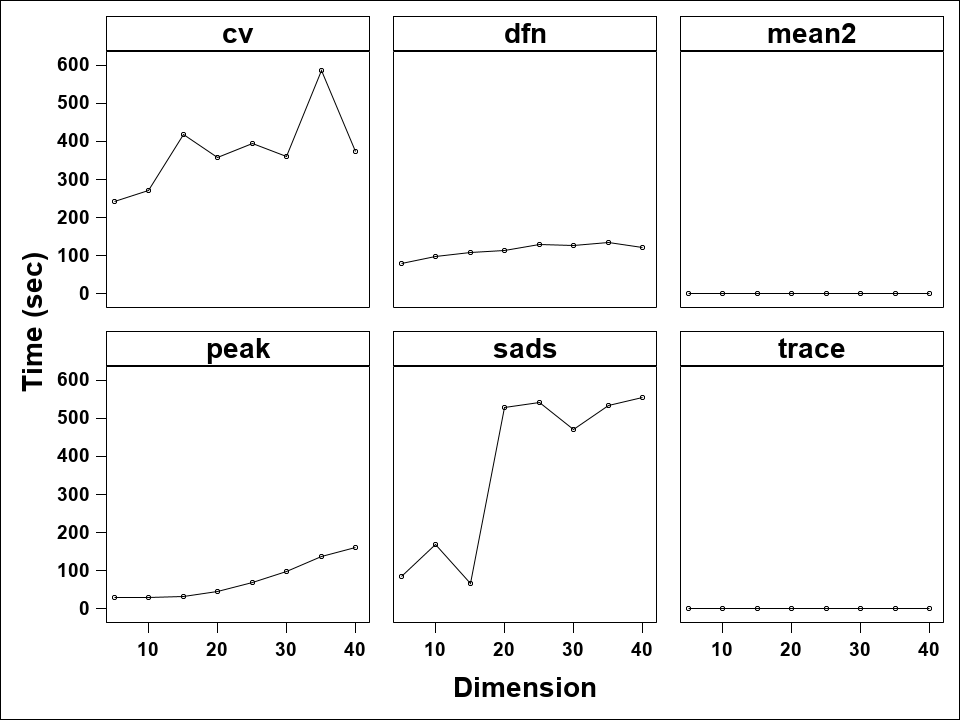}
        }
    \end{tabular}
\end{adjustbox}
\end{table}

\begin{table}
	\centering	
    \captionof{figure}{Simulation results using 10 hypercubes.  CV fails to converge to a bandwidth value for $60\%$ of hypercubes of dimension 5 or greater. SADS fails to converge for $100\%$ of hypercubes of dimension 25 or greater.}\label{fig:multiCubes10}
    \begin{tabular}{cc}
	\subfloat[Evaluation using 10 hypercubes.]{ 
		\includegraphics[width=0.5\textwidth]{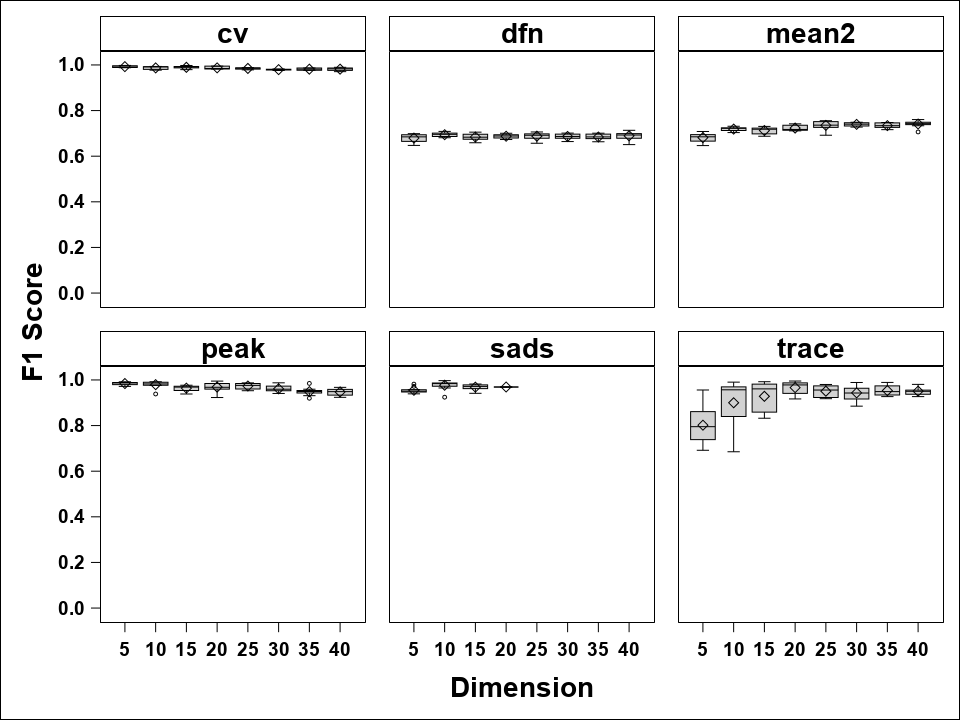}
	}
    &
	\subfloat[Timing using 10 hypercubes]{
		\includegraphics[width=0.5\textwidth]{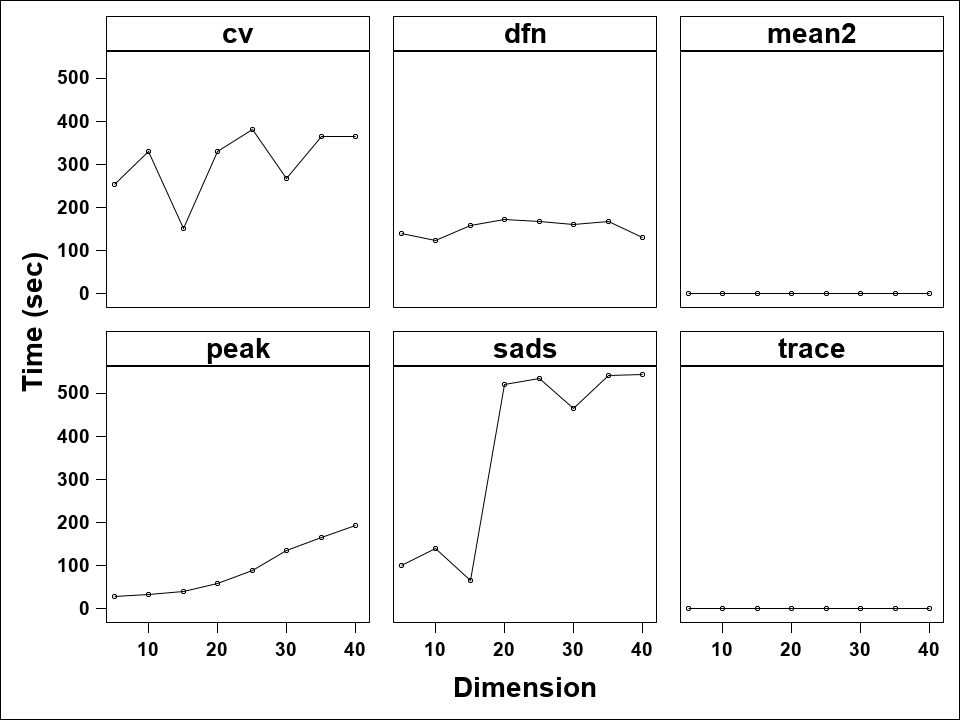}
	}
\end{tabular}
\end{table}

Figure \ref{fig:multiCubes5} and Figure \ref{fig:multiCubes10} provide the box-and-whiskers plot of the $F_1$ measure for simulations results that were obtained using 5 and 10 hypercubes respectively. The CV, peak and trace criterion have the best accuracy performance in this simulation setting. However, CV frequently fails to converge to a bandwidth. SADS does not provide a bandwidth for dimensions greater than 20. Additionally, of th three methods with best accuracy performance, the trace criteria has the best timing performance.

\subsubsection{Evaluation using Increasing Number of Training Observations}
We evaluated the performance of the hypersphere, hypercube, multi-hypersphere and multi-hypercube simulations as the number of training observations $N$ increases. We varied $N$ between $5000$ and $10000$ in increments of $5000$. For each simulation and for each value of $N$, we generated 5 different sets of training and scoring data sets using different seed values. 

\begin{table}[h!]
    \captionof{figure}{Timing results for high-dimensional simulated data}
    \label{tab:incNtrainTab}	
    \begin{adjustbox}{width=0.9\columnwidth,center}
        \begin{tabular}{cc}
            \subfloat[Timing using hyperspheres with dim 20]{
                \includegraphics[width=0.45\textwidth]{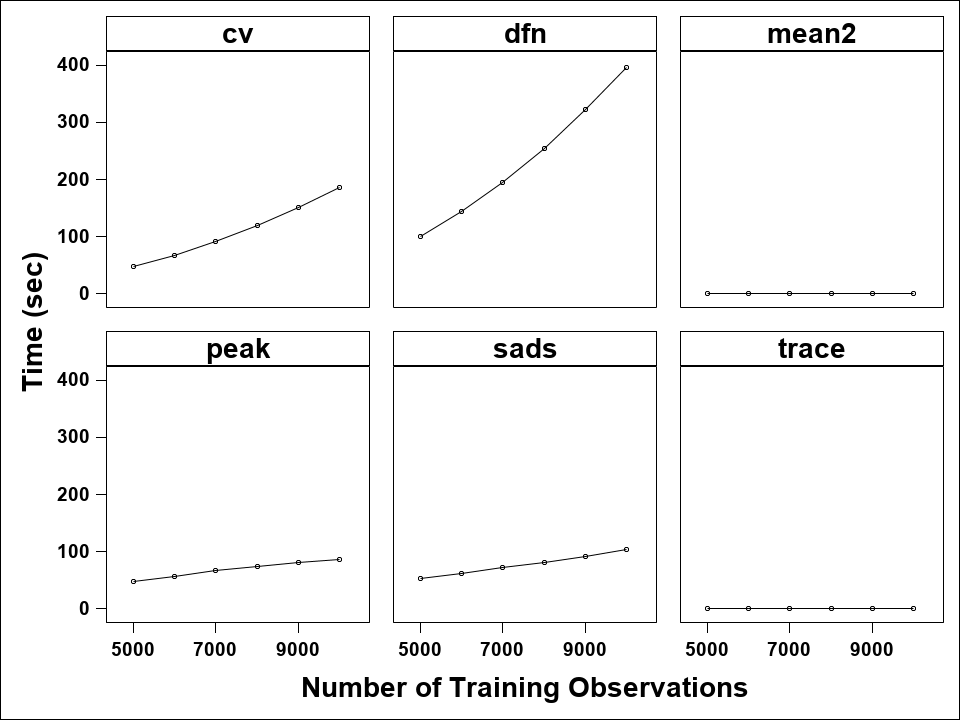}
            } &
            \hfill
            \subfloat[Timing using hypercubes with dim 20]{
                \includegraphics[width=0.45\textwidth]{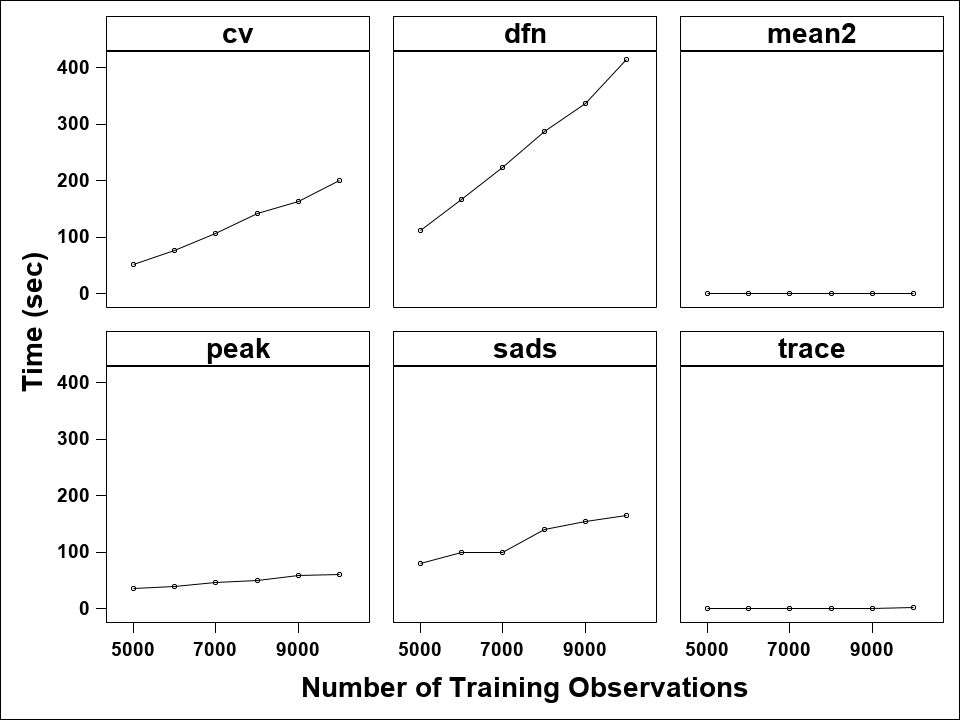}
            } \\
            \hfill
            \subfloat[Timing using 5 hyperspheres with dim 20]{
                \includegraphics[width=0.45\textwidth]{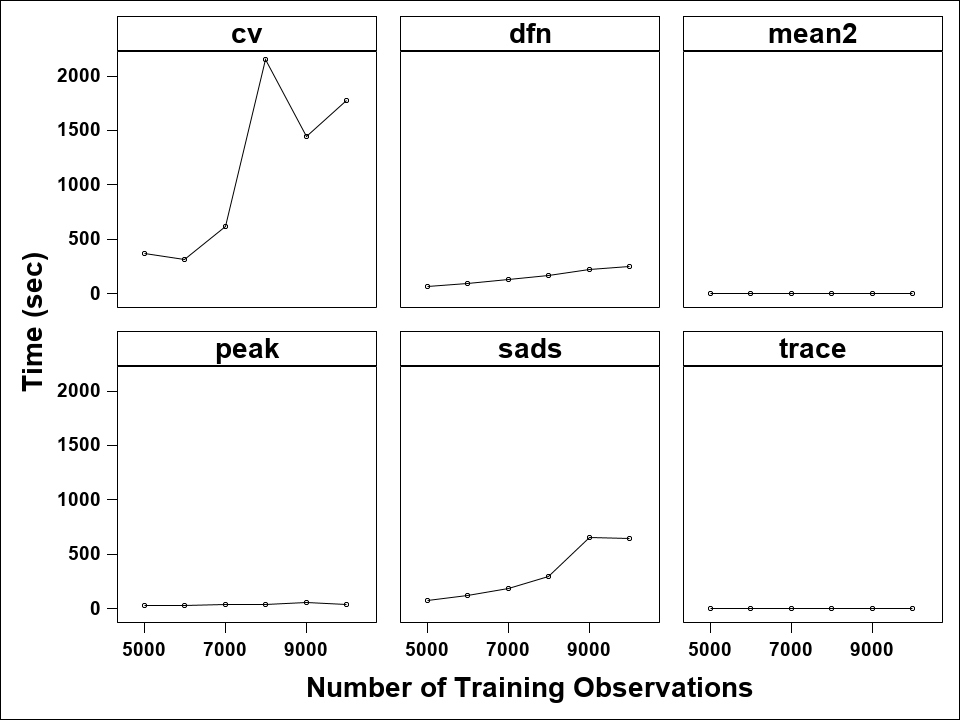}
            } &
            \hfill
            \subfloat[Timing using 5 hypercubes with dim 20]{
                \includegraphics[width=0.45\textwidth]{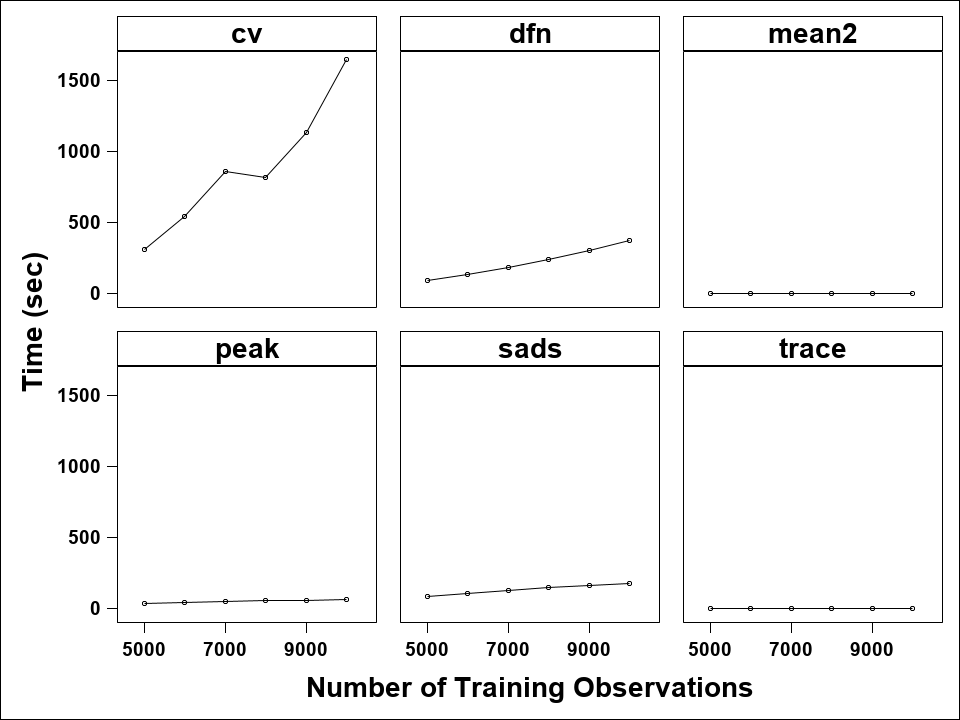}
            } \\
        \end{tabular}
    \end{adjustbox}
\end{table}

The $F_1$ scores corresponding with each bandwidth criteria remained largely
unchanged as $N$ increased with the exception of CV and SADS, which failed to
converge to a bandwidth value for approximately $50\%$ of the data sets with
$N>5000$. Figure \ref{tab:incNtrainTab} shows that for all four simulations, the
modified mean and trace criteria outperform CV, DFN, peak and SADS as the number
of training observations increases both in terms of quality of results and the
time taken for computation.

\section{Conclusion}
The trace criterion for computing the bandwidth value of a Gaussian kernel
for SVDD, as proposed in this paper, exploits the low-rank representation of
the kernel matrix to suggest a bandwidth value. Several evaluations that use
synthetic and real-life data sets indicate that the bandwidth value that is
obtained using the trace criterion provides similar or better results compared
to existing methods and it can be computed quickly even for large datasets. 
In particular the trace criterion method provides the best results when data are
high-dimensional and disjoint.

\section*{Acknowledgement}
Authors would like to thank Anne Baxter, Principal Technical Editor at SAS, for her assistance in creating this
manuscript.

\newpage
\bibliographystyle{acm}
\bibliography{svdd_trace}
\end{document}